\documentclass[journal,comsoc]{IEEEtran}

\usepackage[T1]{fontenc}
\usepackage{graphicx}
\usepackage{xspace}
\usepackage{algorithm}
\usepackage{algpseudocode}
\usepackage{multirow}
\usepackage{arydshln}
\usepackage{xcolor}
\usepackage{subfigure}
\graphicspath{{./figures/}}
\makeatletter
\DeclareRobustCommand\onedot{\futurelet\@let@token\@onedot}
\def\@onedot{\ifx\@let@token.\else.\null\fi\xspace}

\def\eg{\emph{e.g}\onedot}
\def\ie{\emph{i.e}\onedot}

\def\wrt{\emph{w.r.t}\onedot}

\def\etal{\emph{et al}\onedot}
\makeatother

\makeatletter
\newcommand*\bigcdot{\mathpalette\bigcdot@{.5}}
\newcommand*\bigcdot@[2]{\mathbin{\vcenter{\hbox{\scalebox{#2}{$\m@th#1\bullet$}}}}}
\makeatother
\newcommand{\Std}{\operatorname{Std}}
\newcommand{\Expect}{\operatorname{E}}

\ifCLASSINFOpdf
\else
\fi

\usepackage{amsmath}
\usepackage[cmintegrals]{newtxmath}

\begin{document}

\title{SDGMNet: Statistic-based Dynamic Gradient Modulation for Local Descriptor Learning}

\author{Jiayi Ma and Yuxin Deng
\thanks{This work was supported by the National Natural Science Foundation of China under Grant no. 61773295.}
\IEEEcompsocitemizethanks{\IEEEcompsocthanksitem The authors are with the Electronic Information School, Wuhan University, Wuhan, 430072, China (e-mail: jyma2010@gmail.com, dyx\_acuo@whu.edu.cn).}
}

\markboth{}%
{Shell \MakeLowercase{\textit{et al.}}: SDGMNet: Statistic-based Dynamic Gradient Modulation for Local Descriptor Learning}

\maketitle

\begin{abstract}
Modifications on triplet loss that rescale the back-propagated gradients of special pairs have made significant progress on local descriptor learning. However, current gradient modulation strategies are mainly static so that they would suffer from changes of training phases or datasets. In this paper, we propose a dynamic gradient modulation, named SDGMNet, to improve triplet loss for local descriptor learning. The core of our method is formulating modulation functions with statistical characteristics which are estimated dynamically. Firstly, we perform deep analysis on back propagation of general triplet-based loss and introduce included angle for distance measure. On this basis, auto-focus modulation is employed to moderate the impact of statistically uncommon individual pairs in stochastic gradient descent optimization; probabilistic margin cuts off the gradients of proportional Siamese pairs that are believed to reach the optimum; power adjustment balances the total weights of negative pairs and positive pairs. Extensive experiments demonstrate that our novel descriptor surpasses previous state-of-the-arts on standard benchmarks including patch verification, matching and retrieval tasks.
\end{abstract}


\IEEEpeerreviewmaketitle

\section{Introduction}
\IEEEPARstart{E}{valuating} local correspondences of images is a fundamental problem in many computer vision tasks, such as visual localization~\cite{sattler2018benchmarking}, image registration~\cite{ma2014robust} and retrieval~\cite{ng2020solar}. For this purpose, the classic two-stage pipeline including keypoint detection and patch description was proposed decades ago. Although promising end-to-end methods~\cite{yi2016lift,detone2018superpoint,revaud2019r2d2,tian2020d2d,luo2020aslfeat} spring up in recent years, the traditional pipeline remains competitive due to its robustness and efficiency in practice~\cite{ma2020image}. Moreover, deep local descriptors~\cite{han2015matchnet,simo2015discriminative,balntas2016learning,tian2017l2,zhang2017learning,he2018local,keller2018learning,luo2018geodesc,wei2018kernelized,zhang2019learning,tian2019sosnet,wang2019better,ebel2019beyond,tian2020hynet} that are learned with deep neural networks have noticeably outperformed their hand-crafted counterparts,~\eg, SIFT~\cite{lowe2004distinctive}, and boosted the performance of the two-stage pipeline. Therefore, local descriptors based on deep learning merit deeper study.

Benefiting from the great potentials of Deep Neural Network (DNN), deep local descriptors dispense with heuristic designs to acquire invariance as early efforts~\cite{lowe2004distinctive,rublee2011orb} did. Overall, local descriptor learning is exactly a branch of metric learning~\cite{musgrave2020metric}. Specifically, this task aims to encode local patches into discriminative descriptors, and then predict whether pairs of patches are matching or not according to distances between descriptors. To train the encoder, we need to minimize the distance of matching/positive pairs and maximize non-matching/negative ones in the loss function. Various loss functions taking pairs as basic units are proposed, such as softmax loss with deep metric~\cite{han2015matchnet}, pair-wise loss~\cite{kumar2016learning}, triplet loss~\cite{balntas2016learning,mishchuk2017working}, n-pair loss~\cite{tian2017l2} and ranking loss~\cite{he2018local}. Particularly, HardNet~\cite{mishchuk2017working} constructs a potent triplet loss by mining hard negatives in L2Net~\cite{tian2017l2} batch. Recent works mainly focus on modifying this loss due to its simplicity and superiority.

\begin{figure}
	\centering
	\includegraphics[width=0.99\linewidth]{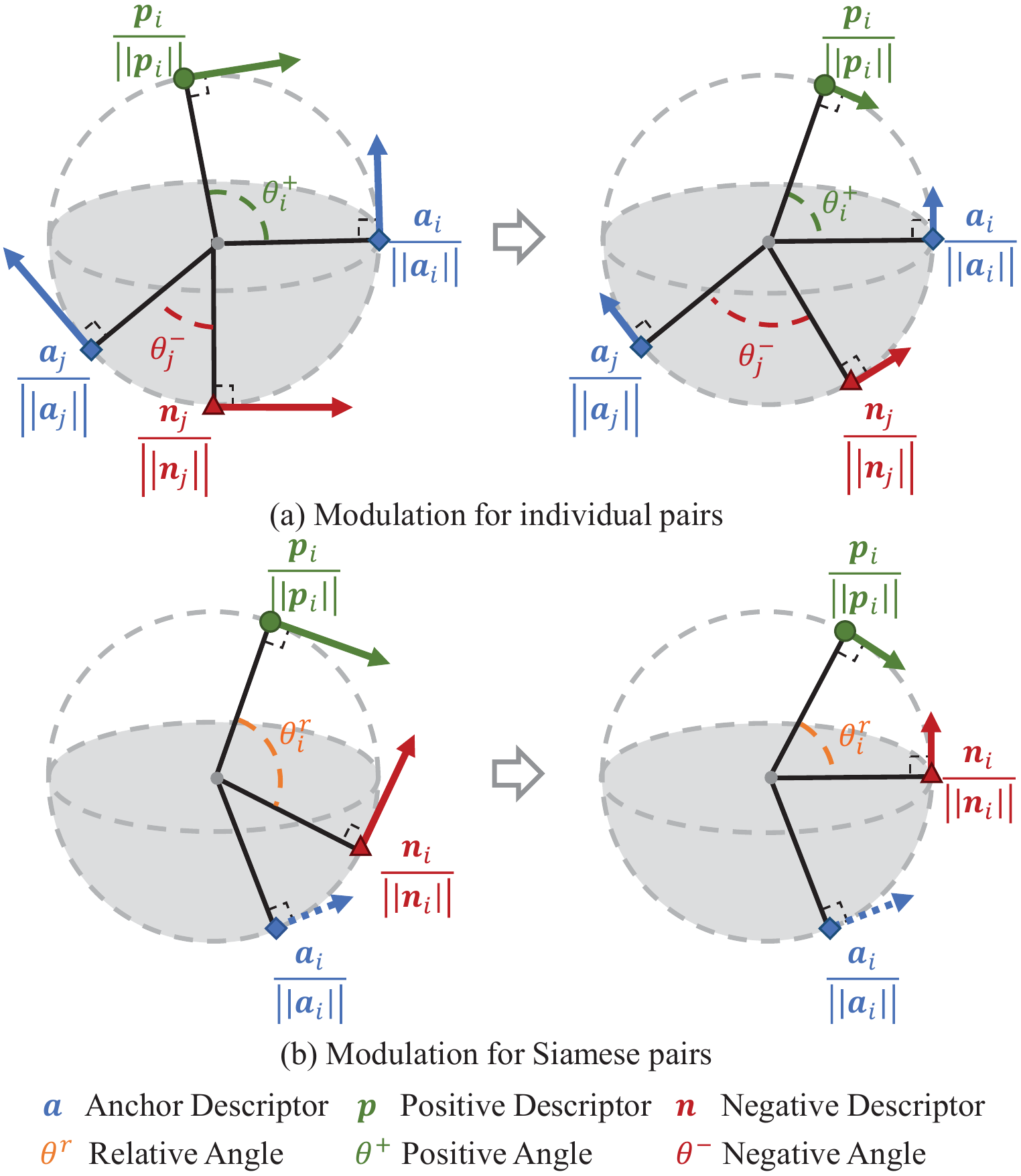}
	\caption{Illustration of the necessity of gradient modulation. Arrows denote the gradients of descriptors during training. And subscripts represent the indices in a batch. Angle for a pair is defined in Eqn.~\eqref{eqn:eqn3}. Relative angle $\theta_i^r$ is equal to $\theta^+_i-\theta_i^-$. (a) The magnitude of the gradient of an individual matching pair $\{\boldsymbol{a}\text{,}\boldsymbol{p}\}$ should increase with $\theta^+$. In contrast, the one of a non-matching pair $\{\boldsymbol{a}\text{,}\boldsymbol{n}\}$ should decrease with $\theta^-$. (b) Siamese pairs,~\ie, a triplet $\{\boldsymbol{a}\text{,}\boldsymbol{p}\text{,}\boldsymbol{n}\}$ in SDGMNet desire a smaller weight, when relative angle $\theta^r$ is diminishing. }
	\label{fig:fig1}
\end{figure}

Besides imposing extra regularization~\cite{zhang2017learning,tian2019sosnet} or resampling patches~\cite{ebel2019beyond}, most modifications~\cite{keller2018learning,wei2018kernelized,zhang2019learning,wang2019better,tian2020hynet} devote to modulating gradients of pairs according to their hardness of discriminating or identifying. Specifically, if an individual positive pair is too distant to identify, its back-propagated gradients should be weighted more largely during optimization. In contrast, the weight for a hard negative pair that is closer should be larger. Moreover, the hardness of Siamese pairs that share the same anchor also deserves attentions. This kind of hardness can be similarly defined with relative distance. These principles for gradient modulation are so-called hard example mining (HEM). We briefly illustrate them in Fig.~\ref{fig:fig1}. HEM also dominates the design of pair-based loss functions in other metric learning tasks~\cite{oh2016deep,wang2017deep,liu2017sphereface,wang2018cosface,deng2019arcface,wang2019multi,sun2020circle,huang2020}. Their successes demonstrate the significance of gradient modulation. However, most modulations are static. The values of the modulation functions depend on individual pairs or Siamese pairs, but do not involve the training phase or the global information. Such modulations suffer from changes of training phases and datasets. Thus, standing on modulating the gradients of individual pairs and Siamese pairs, we concentrate on proposing a dynamic modulation for local descriptor learning. Drawing on global statistical characteristics that are dynamically estimated is promising. Statistics can provide global information that varies over time and datasets, which makes the learning adaptive.

While we are formulating a statistic-based dynamic gradient modulation, more details should be explored. Firstly, related works~\cite{tian2020hynet,zhong2021sface} indicate that once a specific metric,~\eg, $L_2$ distance or inner product is chosen, an implicit modulation harbored in deep back propagation would disorder the elaborate schedule. Secondly, strict HEM should not be encouraged. Because this task is often featured with open-set~\cite{geng2020recent}, few-shot~\cite{wang2020generalizing} and large-scale~\cite{liu2019large}. For example, the training set of \emph{Liberty} in UBC PhotoTour~\cite{winder2007learning} contains $161$K classes with only less than $3$ patches in a class on average. It is unreasonable and impossible to fit the hard pairs perfectly. Finally, the balance between the total modulation weights of negative pairs and positive pairs is fatal for optimization. An overwhelmed ratio of two total weights caused by the modulation would break down the training. In contrast, an appropriate ratio would help convergence and improve generalization.

In this paper, we propose SDGMNet, a statistic-based dynamic gradient modulation to tackle problems mentioned above. SDGMNet is also based on triplet loss proposed by HardNet. We integrate four modifications in SDGMNet. Firstly, we analyze the back-propagated gradient of triplet loss. We explore that angular distance provides the relative flatten magnitude characteristic of gradients before modulation. SDGMNet is easily implemented by pair weighting with the included angle as distance measure (Section~\ref{section3.1}). Secondly, we propose auto-focus modulation to modulate gradients for individual pairs. Auto-focus modulation utilizes the statistical characteristics of distances between individual pairs. It does not follow HEM principle. It mines statistically reliable pairs whose distances lay around the location of the distribution to orient the optimization (Section~\ref{section3.2}). Thirdly, probabilistic margin employs statistical characteristics of relative distance of Siamese pairs,~\ie, triplets. It is applied to cut off the gradients of proportional Siamese pairs that are believed to reach the optimums. Meanwhile, the novel margin draws on hard mining for better convergence (Section~\ref{section3.3}). Finally, we adjust the ratio of positive and negative total weights with weight normalization and attenuation coefficient (Section~\ref{section3.4}). All statistics are estimated dynamically with rough Bayesian sequential update~\cite{bishop2006pattern}. Extensive experiments on standard benchmarks including patch verification, matching, retrieval tasks confirm the superiority of the descriptors learned with SDGMNet.

Our contributions can be summarized as follows:
\begin{itemize}
	\item[1)] We explore the special characteristic of angular distance in back propagation.
	\item[2)] We propose statistics-based auto-focus modulation to moderates the impact of uncommon individual pairs.
	\item[3)] We propose dynamic probabilistic margin to mine Siamese pairs that are hard to discriminate.
	\item[4)] We propose power adjustment to balance total weights of negative and positive pairs for better generalization.
\end{itemize}

\section{Related Works}
In current years, modulating gradients becomes the shared theme of designing pair-based loss functions in metric learning. Modulation strategies can be categorized into two classes: Modulation for individual pairs and modulation for Siamese pairs. We briefly illustrate these two cases with triplet-based loss as an example in Fig.~\ref{fig:fig1}. Our method also stands on them. Thus, we try to absorb related modulation strategies no matter whether they are dynamic or exclusive for triplet loss. For better analyzing these cases in related works, let us represent the general loss as $\mathcal{L}$. $d^+(\boldsymbol{a}\text{,}\boldsymbol{p})$ denotes the general distance between a matching pair $\{\boldsymbol{a}\text{,}\boldsymbol{p}\}$, and $d^-(\boldsymbol{a}\text{,}\boldsymbol{n})$ denotes the one between non-matching $\{\boldsymbol{a}\text{,}\boldsymbol{n}\}$. For convenience, we simplify $d(\ \bigcdot\ )$ as $d$. $d^+-d^-$ is usually referred to relative distance, denoted by $d^r$. $\theta$ is an instance for $d$. Note that, distance and negative similarity are not distinguished deliberately in this paper.

\textbf{Modulation for individual pairs}. Several loss functions, such as Lifted Structure Loss~\cite{oh2016deep}, Binomial Deviance Loss~\cite{yi2014deep} and NCA~\cite{movshovitz2017no}, treat individual positive and negative pairs unequally. However, not all of them are aware of that while $\partial\mathcal{L}/\partial d^+$ should be modulated with an increasing function in terms of $d^+$, $-\partial\mathcal{L}/\partial d^-$ needs decreasing one, if we follow HEM. For local descriptor learning, Keller \etal\cite{keller2018learning} follow HEM. They modulate $\partial\mathcal{L}/\partial d_i^+$ and $-\partial\mathcal{L}/\partial d_i^-$ with functions symmetrical about $(d^+_i+d^-_i)/2$ for each triplet. And then global information is crudely fused by shifting the axis of symmetry. Exp-TL~\cite{wang2019better} conducts more powerful HEM with exponential loss. The new loss makes $\partial\mathcal{L}/\partial d^+$ increase and $-\partial\mathcal{L}/\partial d^-$ decease exponentially. HyNet~\cite{tian2020hynet} observes a hidden modulation in deep back propagation. It substitutes hybrid similarity for common similarity, so the hidden modulation is recast. The new modulation balances the needs of two kinds of pairs. In other tasks, Multi-Simi Loss~\cite{wang2019multi} combines Lifted Structure Loss and Binomial Deviance Loss to satisfy respective demands of $d^+$ and $d^-$. Circle Loss~\cite{sun2020circle} fulfills this purpose with Circle Margin. SFace~\cite{zhong2021sface} employs sigmoid functions to mine hard pairs. Meanwhile, SFace also finds that the functions are disturbed in deep back propagation. But this problem is left alone and defended to restrain the noise in datasets,~\eg, MS-Celeb-1M~\cite{guo2016ms}.

\textbf{Modulation for Siamese pairs}. The relative hardness of Siamese pairs should be also taken into account. It provides more reliable information about the data distribution near the shared anchor. For a triplet, its hardness can be measured by $d_r$. Harder triplets with larger $d_r$ should be emphasized with larger $\partial\mathcal{L}/\partial d^r$ during stochastic gradient descent. Balntas \etal~\cite{balntas2016learning} introduce a static hard margin for local descriptor learning. The hard margin modulates $\partial\mathcal{L}/\partial d^r$ with step function in terms of $d^r$ and prevents easy triplets from backward propagation. Additionally, quadratic triplet loss in SOSNet~\cite{tian2019sosnet} and scale-aware negative logarithmic softmax loss introduced by Keller \etal~\cite{keller2018learning} polish original triplet loss with `soft margin'. Thus, $\partial\mathcal{L}/\partial d^r$ is modulated by continuous functions that monotonically increase with $d^r$. Zhang and Rusinkiewicz~\cite{zhang2019learning} further enroll Cumulative Distribution Function (CDF) to formulate a dynamic soft margin. Furthermore, n-pair losses are more popular with other metric learning tasks. In those cases, angular margin~\cite{liu2017sphereface,wang2018cosface,deng2019arcface} cuts off partial easy Siamese pairs and achieves great success. Multi-Simi~\cite{wang2019multi} Loss separates Siamese pairs into independent positive and negative parts. The gradients of positive or negative pairs that share the same anchor are associated to be modulated. In contrast, Circle Loss~\cite{sun2020circle} considers two kinds of Siamese pairs together.

Although not all of modulations discussed above are customized for triplet loss we try to modify, there is still something worthy of consideration. Especially, Zhang and Rusinkiewicz~\cite{zhang2019learning} have proposed a dynamic modulation for Siamese pairs with CDF, but it is not considerate. Moreover, the deep analysis on back-propagated gradients performed by HyNet~\cite{tian2020hynet} and SFace~\cite{zhong2021sface} suggests that the hidden modulation deserves more attentions. Additionally, the SFace points out that the strict HEM might be irrational, which indirectly explains the nature of hybrid similarity in HyNet. Adsorbing these ideas, we concentrate on our concerns and propose SDGMNet.

\section{Methodology}
Since SDGMNet majors in dynamically modulating gradients of pairs in triplet loss, a deep investigation on back-propagated gradients should be conducted. We define $\boldsymbol{x}(\boldsymbol{\Omega})$ and $\boldsymbol{y}(\boldsymbol{\Omega})$ as descriptors before normalization, where we omit the input patch and keep parameters of encoder $\boldsymbol{\Omega}$ as the only input. To facilitate subsequent analysis, we also discard $(\boldsymbol{\Omega})$. $\boldsymbol{a}$, $\boldsymbol{p}$ and $\boldsymbol{n}$ are instances of $\boldsymbol{x}$ or $\boldsymbol{y}$. Let $N$ denote batch size, $\boldsymbol{D}$ denote the triplet distance batch $\{d_1^-\text{,}d_2^-\text{,}...\text{,}d_N^-\text{;}d_1^+\text{,}d_2^+\text{,}...\text{,}d_N^+\}$. Given a general function $f(\ \bigcdot\ )$ and a distance batch $\boldsymbol{D}$, the general triplet loss $\mathcal{L}$ can be represented as
\begin{equation}
	\mathcal{L}(\boldsymbol{D})={f(d_1^-,d_2^-,...,d_N^-;d_1^+,d_2^+,...,d_N^+)}\text{.}
	\label{eqn:eqn1}
\end{equation}
The back-propagated gradient of the loss \wrt the parameters of the encoder $\boldsymbol{\Omega}$ can be computed with chain rule~\cite{goodfellow2016deep} as:
\begin{equation}
	\begin{aligned}
		\frac{{\partial \mathcal{L}}}{{\partial \boldsymbol{\Omega}}} = &\sum\limits_{i = 1}^N\frac{\partial\mathcal{L}}{\partial d_i^+}\Big(\frac{{\partial d_i^+ }}{{\partial {\boldsymbol{a}_i}}} \frac{{\partial \boldsymbol{a}_i}}{{\partial \boldsymbol{\Omega}}}+ \frac{{\partial d_i^+ }}{{\partial {\boldsymbol{p}_i}}}\frac{{\partial \boldsymbol{p}_i}}{{\partial \boldsymbol{\Omega}}}\Big)+\\&\sum\limits_{i = 1}^N\frac{\partial\mathcal{L}}{\partial d_i^-}\Big(\frac{{\partial d_i^- }}{{\partial {\boldsymbol{a}_i}}} \frac{{\partial \boldsymbol{a}_i}}{{\partial \boldsymbol{\Omega}}}+ \frac{{\partial d_i^- }}{{\partial {\boldsymbol{n}_i}}}\frac{{\partial \boldsymbol{n}_i}}{{\partial \boldsymbol{\Omega}}}\Big)\text{,}
	\end{aligned}
	\label{eqn:eqn2}
\end{equation}
where $(\boldsymbol{D})$ is omitted. In Eqn.~\eqref{eqn:eqn2}, $\partial\mathcal{L}/\partial d$ is a scalar. It reveals how much the corresponding pair contributes to update of parameters. Gradient modulation for pairs focuses on rescaling $\partial\mathcal{L}/\partial d$ with a function about $d$. In this way, stochastic gradient descent optimization can be better controlled. However, related works~\cite{zhong2021sface,tian2020hynet} explore that the term $\partial d/\partial \boldsymbol{x}$ harbors a hidden modulation function about $d$ which would break our intention.

\subsection{Angular Distance}
\label{section3.1}
Consider the term $\partial d/\partial \boldsymbol{x}$. Due to normalization, descriptors are embedded onto unit hypersphere. Included angular $\theta$ can be used for distance measure. It is defined as
\begin{equation}
	\label{eqn:eqn3}
	\theta=\text{acos}\frac{\boldsymbol{x}\boldsymbol{y}}{\|\boldsymbol{x}\|\|\boldsymbol{y}\|}\text{,}
\end{equation}
where $\|\bigcdot\|$ denotes $L_2$ normalization. There are two more common metrics for distance measure: inner product $s$ and $L_2$ distance $l$. They are defined as
\begin{equation}
	\label{eqn:eqn4}
	s=\frac{\boldsymbol{x}\boldsymbol{y}}{\|\boldsymbol{x}\|\|\boldsymbol{y}\|}\text{,}\qquad l=\Big\|\frac{\boldsymbol{x}}{\|\boldsymbol{x}\|}-\frac{\boldsymbol{y}}{\|\boldsymbol{y}\|}\Big\|\text{.}
\end{equation}
They are equivalent instances of $d$ in forward propagation but distinguishing in back propagation. In back propagation, $\partial\theta/\partial \boldsymbol{x}$, $\partial s/\partial \boldsymbol{x}$ and $\partial l/\partial \boldsymbol{x}$ share the same optimal direction which is orthogonal to $\boldsymbol{x}$ as illustrated in Fig.~\ref{fig:fig1}. However, they own special magnitudes as
\begin{align}
	\label{eqn:eqn5}
	\Big\|\frac{\partial \theta}{\partial \boldsymbol{x}}\Big\|&=\frac{1}{\|\boldsymbol{x}\|}\text{,}\\
	\label{eqn:eqn6}
	\Big\|\frac{\partial s}{\partial \boldsymbol{x}}\Big\|&=\frac{1}{\|\boldsymbol{x}\|}\sqrt{1-s^2}\text{,}\\
	\label{eqn:eqn7}
	\Big\|\frac{\partial l}{\partial 	\boldsymbol{x}}\Big\|&=\frac{1}{\|\boldsymbol{x}\|}\frac{\sqrt{4l^2-l^4}}{2l}\text{.}
\end{align}

As shown above, an implicit modulation takes effect after a metric is chosen. The magnitude of $\partial\theta/\partial \boldsymbol{x}$ depends on $\|\boldsymbol{x}\|$ only, which would not disturb the modulation function about $\theta$ we design later. Thus, $\theta$ is a suitable choice for our further intention. As for the term $1/\|\boldsymbol{x}\|$, we free it for two reasons. Firstly, Ranjan \etal~\cite{ranjan2017l2} observe that a high-quality picture tends to carry large magnitude of feature in similar practice. According to Eqn.~\eqref{eqn:eqn5}, high-quality pictures will receive weaker updates and support the feature space at the latter period of training, which may facilitate the optimization. Moreover, $\boldsymbol{a}$, $\boldsymbol{p}$ and $\boldsymbol{n}$ interchange in different batches. So the impact of $1/\|\boldsymbol{x}\|$ holds balanced globally.

In short, $\partial \theta / \partial \boldsymbol{x}$ owns the optimal direction and a plain magnitude for learning. Thus, we employ $\theta$ for distance measure in SDGMNet. As a result, we can dedicate to gradient modulation for pairs,~\ie, formulating $\partial\mathcal{L}/\partial d$. Eqn.~\eqref{eqn:eqn2} can be reformulated with $\theta$ into
\begin{equation}
	\frac{{\partial \mathcal{L}}}{{\partial \boldsymbol{\Omega}}} = \sum\limits_{i = 1}^N{w^+_i\frac{{\partial \theta_i^+}}{{\partial {\boldsymbol{\Omega}}}}}-\sum\limits_{i= 1}^N{w^-_i\frac{{\partial \theta_i^-}}{{\partial {\boldsymbol{\Omega}}}}}\text{,}
	\label{eqn:eqn8}
\end{equation}
where $\partial\mathcal{L}/\partial d_i$ is replaced by a weight $w_i$ for convenience. We decompose $w_i$ into $w_s\times w_c$ in SDGMNet. $w_s$ and $w_c$ will be introduced in following subsections.

\begin{figure}
	\centering
	\includegraphics[width=1.0\linewidth]{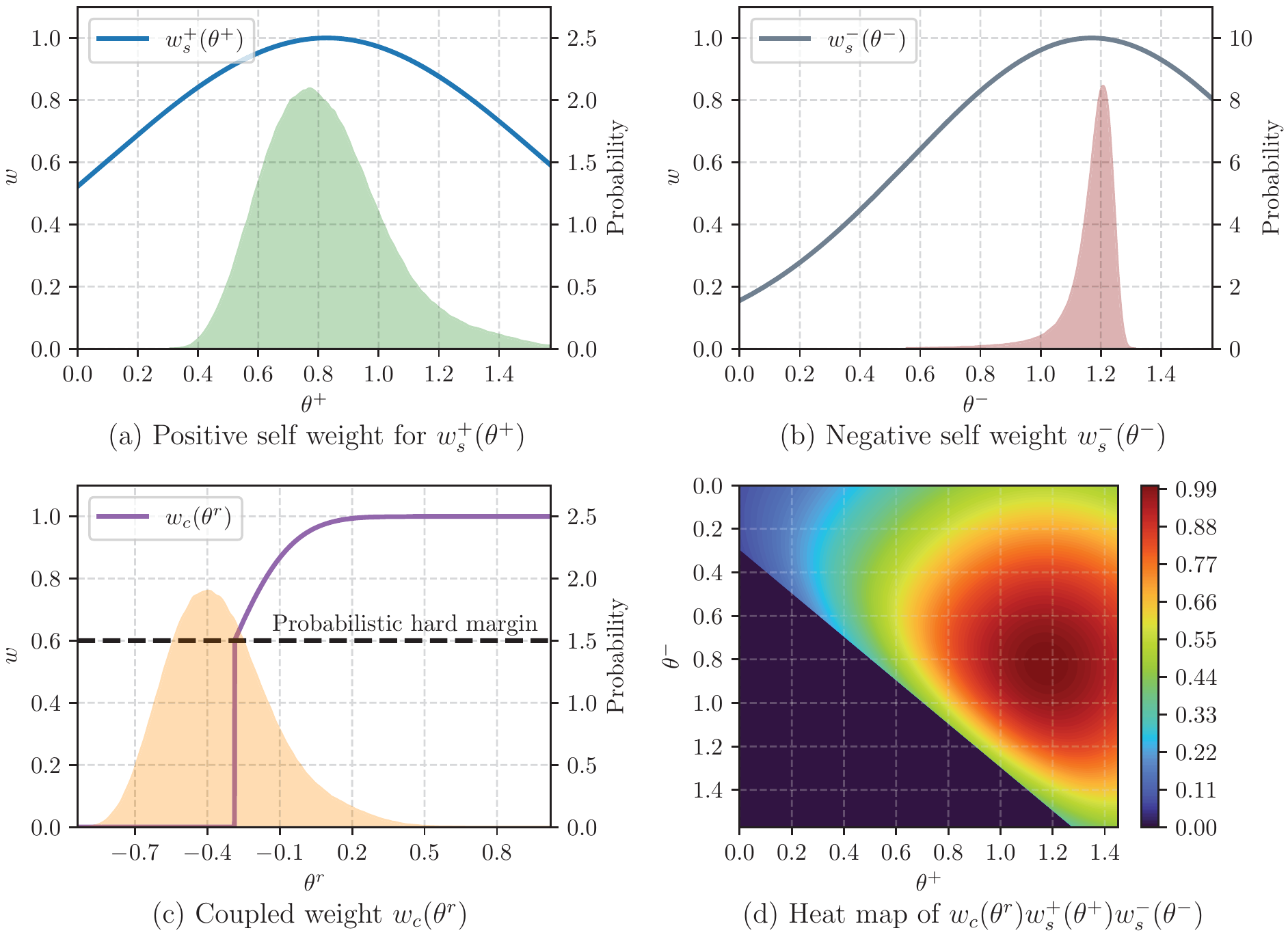}
	\caption{Visualization of modulation functions and related data distributions at the last training epoch on \emph{Liberty}. Curves in (a), (b) and (c) illustrate three kinds of weights introduced in the text. Shadows demonstrate the probability distribution of variables. (d) is a heat map of $w_s^+(\theta^+)w_s^-(\theta^-)w_c(\theta^r)$ with $\theta^+$ and $\theta^-$ as x-axis and y-axis. The dark red in (d) indicates the strong impact on optimization. Our formulation does not turn the spotlight on hardest triplets that should lay at the top right corner. Easy triplets at the bottom left corner are eliminated by the hard margin.}\label{fig:fig2}
\end{figure}

\subsection{Auto-focus Modulation}
\label{section3.2}
$w_s$ is the weight for modulating the gradients of individual pairs. $w_s$ is refereed to self weight because it is exclusive for a pair.

Ideally, the angular distances of individual negative and the positive pairs reach their own optimums at $\pi$ and $0$, respectively. If following HEM, the gradient of an angle that is further away from its optimum should be weighted more largely. In other words, the gradients of matching pairs should be modulated with $w_s^+$ that is monotonously increasing~\wrt $\theta^+$, and the non-matching with monotonously decreasing $w_s^-$. However, naive distances $\theta^+$ and $\theta^-$ are not dependable. Although the hardest positive pairs with large $\theta^+$ are collected correctly, extreme distortions they carry would damage the convergence of stochastic gradient descent optimization. For the hardest negative pairs of patches, while the real distance between them cannot be evaluated, we simply push their descriptors away. Thus, extreme individual pairs should be treated more cautiously. Successes of HyNet~\cite{tian2020hynet} and Sface~\cite{zhong2021sface} also imply that excessive HEM on individual pairs should not be advocated.

To neutralize the HEM and extreme individual pairs suppression, we proposes auto-focus modulation to formulate dynamic $w_s^+$ and $w_s^-$ in SDGMNet as
\begin{equation}
	w_s^+(\theta^+)=\exp(-\frac{(\theta^+-\Expect_t[\theta^+])^2}{2(\pi/6+\Std_t[\theta^+])^2})\text{,}
	\label{eqn:eqn9}
\end{equation}
\begin{equation}
	w_s^-(\theta^-)=\exp(-\frac{(\theta^--\Expect_t[\theta^-])^2}{2(\pi/6+\Std_t[\theta^-])^2})\text{,}
	\label{eqn:eqn10}
\end{equation}
where $\Expect[\,\bigcdot\,] $ represents the expectation, $\Std[\,\bigcdot\,]$ denotes the standard deviance, and the subscript $t$ means the statistical characteristics are dynamically estimated over time. We visualize $w_s^+(\theta^+)$ and $w_s^-(\theta^-)$ at the last training iteration in Fig.~\ref{fig:fig2} (a) and (b). The auto-focus modulation originates from Gaussian blur. It automatically focuses on the expectation of the positive or negative pairs that are more reliable examples we believe. Meanwhile, the impacts of harder and easier examples are weaken. It is worth mentioning that we limit the lower bound of the blur radius to $\pi/2$,~\ie, add $\pi/6$ to the standard deviance. If no constraint, the hardest examples that are long-tailed will be cleaned out with extremely small weight. Since the angle of positive pairs and negative pairs spread mainly on $[0,\pi/2]$ attribute to curse of dimensionality~\cite{zhang2017learning}, a radius equal to $3(\pi/6+\Std_t(\theta))$ is suitable to cover all examples with considerable weights.

\subsection{Probabilistic Margin}
\label{section3.3}
The stochastic gradient descent optimization does not converge until $\partial \mathcal{L}/\partial \boldsymbol{\Omega}$ approaches zero. Thus, margins are necessary to force $w$ in Eqn.~\eqref{eqn:eqn8} to be zeros near the optimums. For example, $\theta_i^+$ holds an ideal optimum at $0$, so the ${w_i^+}|_{\theta _i< 0+m_i}$ should be 0, where the exclusive margin $m_i$ can be an infinitesimal. However, it is impossible to search for $m_i$ for each pair. Furthermore, two margins for matching and non-matching pairs are rational, but a joint margin set for triplets always proves more productive in practice. In other words, a single margin $m$ designed as a function of $\theta^r$ is favored. The function modulates Siamese pairs with the same weight. The value of the function couples Siamese pairs, so we name it coupled weight $w_c$. How large $m$ should be set for the global optimum is still confused. Instead of employing a fixed empirical $m$, we intend to elaborate a dynamic hard margin that believes $100\times m\%$ examples have reached the optimum, so-called probabilistic hard margin.

Given a probabilistic hard margin $m$, we learn from the CDF-based soft margin~\cite{zhang2019learning} to form $w_c$ in SDGMNet as
\begin{equation}
	w_c(\theta^r)=\left\{
	\begin{aligned}
		\text{CD}&\text{F}_t(\theta^r)\text{,}\quad && \theta^r>\text{iCDF}_t(m)\text{,} \\
		&0\text{,} &&\theta^r\le\text{iCDF}_t(m)\text{,}
	\end{aligned}
	\right.
	\label{eqn:eqn11}
\end{equation}
where CDF$(\,\bigcdot \,)$ denotes cumulative distribution function, iCDF$(\,\bigcdot\,)$ denotes inverse cumulative distribution function. Triplets that carry $\theta^r$ smaller than iCDF$(m)$ are top $100\times m\%$ easy examples empirically. These easy examples are believed to approach the optimum and will be isolated from further optimization. The others are preserved and weighted by a monotonously increasing CDF for HEM. HEM for modulating Siamese pairs works well. Due to probabilistic hardness, the modulation is dynamic and adaptive to training data and stage. To facilitate the implementation, we approximate the data distribution with a normal distribution. The curve of $w_c(\theta^r)$ is drawn in Fig.~\ref{fig:fig2} (c), where we set $m=0.6$. As can be seen, probabilistic hard margin steepens the CDF-based coupled weight to cut off proportional easy triplets.

So far, the coupled weight and self weight in SDGMNet are conceived completely.

\subsection{Power Adjustment}
\label{section3.4}
We define power as the total weight of a class of pairs:
\begin{equation}
	P^+ = \sum_{i=1}^N{w_i^+}\text{,} \qquad P^- = \sum_{i=1}^N{w_i^-}\text{.}
	\label{eqn:eqn12}
\end{equation}
Power describes how strongly a class of positive or negative pairs guides the training with gradients. Before modulation,~\ie, $w=1$, the negative power $P^+$ and the positive power $P^-$ hold balanced. However, customized modulations on two classes make powers out of control. Intuitively, the positive power guides the model to identify the images with the same label. In contrast, the negative power forces the model to discriminate non-matching examples. An inductive bias on the negative power $P^-$ would be preferred. Because the model need not identify all labels well which would not appear in the test. Moreover, discriminating ability mutually promotes identifying for human beings. To adjust the ratio of power, we propose weight normalization that divides the weights by the expectation of the corresponding power. Then, attenuation is adopted on the positive side. Finally, SDGMNet is finished as:
\begin{equation}
	\frac{{\partial \mathcal{L}}}{{\partial \boldsymbol{\Omega}}} = \frac{\alpha}{E_t[P^+]}\sum\limits_{i = 1}^N{w^+_i\frac{{\partial \theta_i^+}}{{\partial {\boldsymbol{\Omega}}}}}-\frac{1}{E_t[P^-]}\sum\limits_{i= 1}^N{w^-_i\frac{{\partial \theta_i^-}}{{\partial {\boldsymbol{\Omega}}}}}\text{,}
	\label{eqn:eqn13}
\end{equation}
where $\alpha$ is the attenuation coefficient.

Without normalization, the scale factor in the previous modulation function would be enabled. Choosing an appropriate scale would be complicated. Moreover, even if a static modulation is employed, the unnormalized power would be still unpredictable due to the varying distribution of data. The ratio of the positive power to the negative power cannot be controlled under such circumstances. Once normalization functions, the ratio of the normalized powers can be quantified and adjusted by the attenuation coefficient. Such a ratio can endure random data, arbitrary modulations and running training phases. Furthermore, the total weight can be regarded as a kind of equivalent learning rate for $\partial \theta /\partial \boldsymbol{\Omega}$. Weight normalization guarantees the equivalent learning rate to be stable so that the real learning rate of the optimizer would not be disturbed. As a result, the learning rate can pace as it is proposed to do.

\begin{figure}
	\centering
	\includegraphics[width=0.9\linewidth]{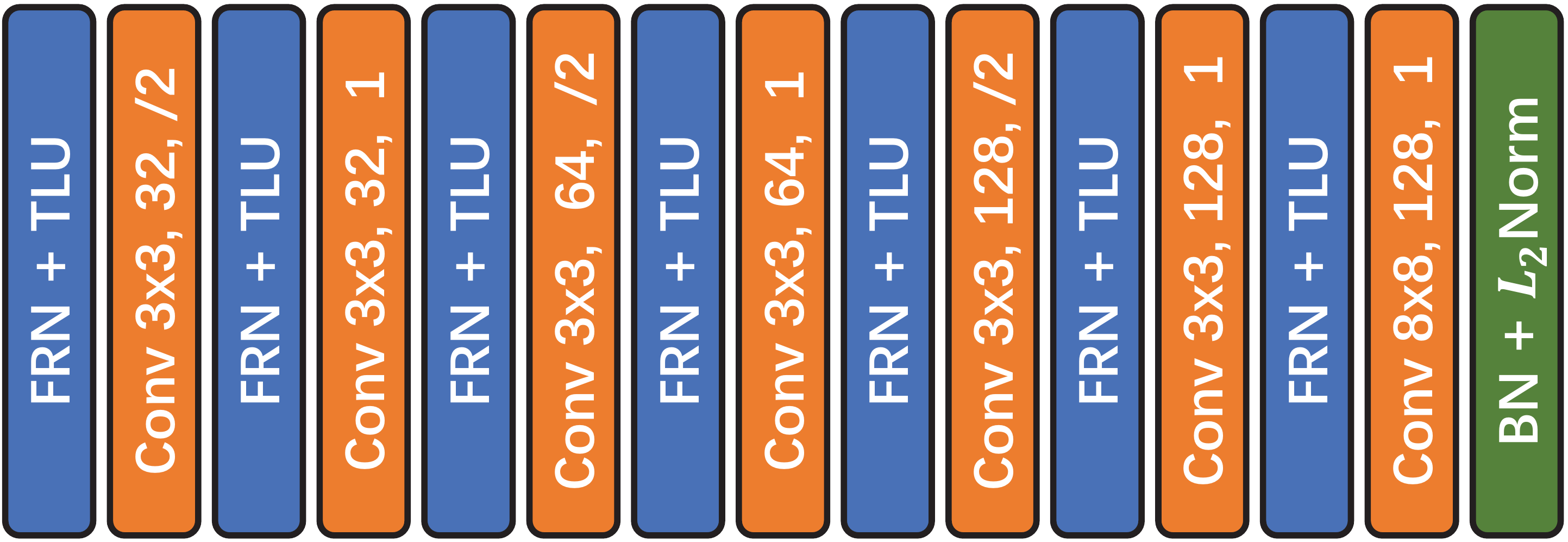}
	\caption{Network architecture adopted from HyNet~\cite{tian2020hynet}. Biases in convolution layers are activated except for the last one. Dropout regularization with 0.3 dropout rate is used before the last convolution layer.}
	\label{fig:fig3}
\end{figure}

\subsection{Implementation}
\label{section3.5}
\textbf{Triplet sampling}. Triplet sampling strategy proposed by HardNet~\cite{movshovitz2017no} has become the de-facto standard for local descriptor learning. Briefly, HardNet follows L2Net~\cite{tian2017l2} to sample $N$ matching pairs. For a matching pair, HardNet mines the nearest negative neighbor as the negative example in the triplet. We follow the strategy and employ the anti-noise threshold for hard negative mining. The threshold in angular distance is set to 0.6.

\textbf{Network architecture}. L2Net~\cite{tian2017l2} proposes a classic encoder, in which batch normalization~\cite{ioffe2015batch} and ReLU are employed after each convolution layer except for the last one. HyNet~\cite{tian2020hynet} replaces these normalization layers to learnable Filter Response Normalization (FRN) and TLU~\cite{singh2020filter}. Moreover, it introduces an additional normalization layer at the input of the encoder. These modifications remarkably improve local deep descriptor learning with little cost. We adopt this network architecture and illustrate it in Fig.~\ref{fig:fig3}.

\textbf{Statistics estimation}. There are some statistical arguments varying over time in SDGMNet, which make the modulation dynamic. We employ rough Bayesian sequential update~\cite{bishop2006pattern} to estimate these arguments as:
\begin{equation}
	\boldsymbol{\beta}{_t} = 0.999\boldsymbol{\beta} _{t-1}+0.001\boldsymbol{\mu}_t\text{,}
	\label{eqn:eqn14}
\end{equation}
where $\boldsymbol{\beta}_t$ is the vector of approximated global statistics and $\boldsymbol{\mu}_t$ is the estimation in batch at $t$th iteration. The past data provide a priori to current training. A fixed replacement rate $0.001$ is sensitive enough for the former stage and stable for the latter stage of training.

\begin{algorithm}[t]
	\caption{SDGMNet for local descriptor learning}
	\label{alg:alg1}
	\hspace*{3.5mm}\textbf{Input:} SDGMNet hyperparameters $m$ and $\alpha$, vector of initial statistics $\boldsymbol{\beta}_0$, raw model, dataset, optimizer.
	
	\hspace*{3.5mm} $t=1$;
	\begin{algorithmic}
		\While{training}
		\State Sample a data batch from datasets;
		\State Compute the angular distance matrix;
		\State Obtain HardNet triplets;
		\State Compute statistics $\boldsymbol{\mu}_t$ in batch, except for the expectation of powers;
		\State Update corresponding statistics by Eqn.~\eqref{eqn:eqn14};
		\State Compute $w_s^+(\theta^+)$, $w_s^-(\theta^-)$ and $w_c(\theta^r)$ by Eqns.~\eqref{eqn:eqn9},~\eqref{eqn:eqn10} and~\eqref{eqn:eqn11} respectively;
		\If{warming}
			\State Set $w_s^+(\theta^+)$, $w_s^-(\theta^-)$ and $w_c(\theta^r)$ to $1$
		\EndIf
		\State Compute powers $P^+$ and $P^-$ by Eqn.~\eqref{eqn:eqn12};
		\State Update the expectation of powers by Eqn.~\eqref{eqn:eqn14};
		\State Construct pseudo loss by Eqn.~\eqref{eqn:eqn15};
		\State Call back propagation of pseudo loss;
		\State Update the model with the optimizer;
		\State $t=t+1$;
		\EndWhile
	\end{algorithmic}
	\hspace*{3.5mm}\textbf{Output:} Well-trained model.
	\label{alg}
\end{algorithm}

\begin{table*}[t]
	\centering
	\setlength{\tabcolsep}{5.0mm}
	\renewcommand\arraystretch{1.3}
	\caption{Patch verification performance on UBC PhotoTour. Numbers shown are FPR@95 that are lower for better. The best scores are highlighted in bold. Dash lines indicate changes of models. HardNet+FRN denotes the version of HardNet trained with SDGMNet settings. LIB: \emph{Liberty}, YOS: \emph{Yosemite}, ND: \emph{Notredame}.}
	\begin{tabular}{c|cccccccc|c}
		\hline
		\multicolumn{1}{c|}{Train} & \multicolumn{1}{c}{ND}   & \multicolumn{1}{c}{YOS}   &                      & \multicolumn{1}{c}{LIB}  & \multicolumn{1}{c}{YOS}  &                      & \multicolumn{1}{c}{LIB}  & \multicolumn{1}{c|}{ND}   & \multirow{2}{*}{Mean} \\ \cline{1-3} \cline{5-6} \cline{8-9}
		Test                       & \multicolumn{2}{c}{LIB}                              &                      & \multicolumn{2}{c}{ND}                              &                      & \multicolumn{2}{c|}{YOS}                              &                       \\ \hline
		SIFT~\cite{lowe2004distinctive}                       &  \multicolumn{2}{c}{29.84}                            &                      & \multicolumn{2}{c}{22.53}                           &                      & \multicolumn{2}{c|}{27.29}                            & 26.55                 \\ \hdashline
		TFeat~\cite{balntas2016learning}                      & \multicolumn{1}{c}{7.39} & \multicolumn{1}{c}{10.13} & \multicolumn{1}{c}{} & \multicolumn{1}{c}{3.06} & \multicolumn{1}{c}{3.80} & \multicolumn{1}{c}{} & \multicolumn{1}{c}{8.06} & \multicolumn{1}{c|}{7.24} & 6.64                  \\ \hdashline
		L2Net~\cite{tian2017l2}                    & 2.36                     & 4.70                      &                      & 0.72                     & 1.29                     &                      & 2.57                     & 1.17                      & 2.23                  \\
		HardNet~\cite{mishchuk2017working}                    & 1.49                     & 2.51                      &                      & 0.53                     & 0.78                     &                      & 1.96                     & 1.84                      & 1.51                  \\
		CDFDesc~\cite{zhang2019learning}                        & 1.21                     & 2.01                      &                      & 0.39                     & 0.68                     &                      & 1.51                     & 1.29                      & 1.38                  \\
		SOSNet~\cite{tian2019sosnet}                     & 1.08                     & 2.12                      &                      & 0.34                     & 0.67                     &                      & 1.03                     & 0.95                      & 1.03                  \\ \hdashline
		HardNet+FRN~\cite{mishchuk2017working,tian2020hynet}                      & 1.26                     & 1.76                      &                      & 0.41                     & 0.58                     &                      & 1.16                     & 1.05                     & 1.04                  \\
		HyNet~\cite{tian2020hynet}                      & 0.89                     & \textbf{1.37}                      &                      & 0.34                     & 0.61                     &                      & 0.88                     & 0.96                      & 0.84                  \\
		SDGMNet                     & \textbf{0.82}                    & 1.41                      &                      & \textbf{0.30}                     & \textbf{0.46}                     &                      & \textbf{0.80}                     & \textbf{0.76}                      & \textbf{0.76}                  \\ \hline
	\end{tabular}
	\label{tab:tab1}
\end{table*}

\begin{table}[t]
	\centering
	\setlength{\tabcolsep}{2.5mm}
	\renewcommand\arraystretch{1.3}
	\caption{Statistics on \emph{Liberty}}
	\label{tab:tab2}
	\begin{tabular}{c|ccccc}
		\hline
		Epoch-$th$ & 30    & 70    & 110   & 150   & 190   \\ \hline
		$\Expect[\theta^r](10^{-1})$       & -2.86 & -3.13 & -3.30 & -3.39 & -3.43 \\ \hline
		$\Std[\theta^r](10^{-1})$       & 2.37  & 2.31  & 2.25  & 2.21  & 2.18  \\ \hline
		$\Expect[\theta^+](10^{-1})$        & 8.53  & 8.39  & 8.31  & 8.28  & 8.26  \\ \hline
		$\Std[\theta^+](10^{-1})$       & 2.24  & 2.16  & 2.11  & 2.04  & 2.03  \\ \hline
		$\Expect[\theta^-]$       & 1.14  & 1.15  & 1.16  & 1.16  & 1.17  \\ \hline
		$\Std[\theta^-](10^{-2})$       & 7.78  & 7.97  & 8.09  & 8.15 & 8.14
		\\	\hline	
		$\Expect[P^+]$       & 281  & 280  & 280  & 279 & 279
		\\ \hline
		$\Expect[P^-]$       & 299  & 297  & 295  & 294 & 294
		\\
		\hline
	\end{tabular}
\end{table}

\textbf{Pseudo loss}. We avoid an explicit loss that owns the gradient shown in Eqn.~\eqref{eqn:eqn13}. Motivated by general pair weighting framework~\cite{wang2019multi}, we define pseudo loss as:
\begin{equation}
	\mathcal{L_P} = \frac{\alpha}{E_t[P^+]}\sum\limits_{i = 1}^N{w^+_i\theta^+_i}-\frac{1}{E_t[P^-]}\sum\limits_{i = 1}^N{w^-_i\theta^-_i}\text{,}
	\label{eqn:eqn15}
\end{equation}
where $\alpha$, $w$, $E_t[P^+]$ and $E_t[P^-]$ are all constant with regard to the model parameter $\boldsymbol{\Omega}$. The gradient of pseudo loss is the same as Eqn.~\eqref{eqn:eqn13} so it can be used to exercise the SDGMNet. Pseudo loss holds an ambiguous optimum and does not satisfy the conditions to be a loss. It is a more general tool to guide the training with arbitrary gradient modulation strategies.

\textbf{Settings}. We implement SDGMNet in PyTorch~\cite{paszke2019pytorch}. The procedure of SDGMNet is shown in Algorithm~\ref{alg}, where we set $m=0.6$ and $\alpha=0.9$ for the best performance. The network is trained for $200$ epochs ($200$K iterations) with batch size of 1024 and SGD optimizer. Data augmentation is achieved by random rotation, flipping and cropping~\cite{mishchuk2017working,zhang2019learning,shorten2019survey}. Momentum and weight decay of the optimizer are set to $0.9$ and $0.0001$, respectively. The learning rate is initialized with 1 and divided by 2 after each $10\%$ of iterations. Moreover, the training is warmed up with $w=1$ in the first $10\%$ of iterations. During warming, only $E[P^+]$ and $E[P^-]$ function but all statistics are estimated in every iteration. As a result, only the initial values of $E[P^+]$ and $E[P^-]$ contribute to the full SDGMNet training. We set them to 10000 so that the equivalent learning rate could act similarly as one cycle learning rate policy~\cite{smith2017cyclical}.

\section{Experiments}
We experiment SDGMNet on three benchmarks: UBC PhotoTourism~\cite{winder2007learning}, HPatches~\cite{balntas2017hpatches} and ETH 3D reconstruction~\cite{schonberger2017comparative}. Train settings are shown in Section~\ref{section3.5}. The test results are compared with state-of-the-art alternatives including SIFT~\cite{lowe2004distinctive}, TFeat~\cite{balntas2016learning}, L2Net~\cite{tian2017l2}, HardNet~\cite{mishchuk2017working}, CDFDesc~\cite{zhang2019learning}, SOSNet~\cite{tian2019sosnet} and HyNet~\cite{tian2020hynet}. All methods output 128-dimensional descriptors that can be evaluated with $L_2$ distance. Those learned with DNN are trained with data augmentation. HardNet, CDF and SOSNet enroll the same network architecture proposed by L2Net, which is deeper than TFeat's. HyNet further upgrades the network with FRN. Not that because L2Net, SOSNet and HyNet are not open source, we just report the results from their published papers. For the rest, we validate their records with released codes.

\begin{table}[t]
	\centering
	\setlength{\tabcolsep}{2.5mm}
	\renewcommand\arraystretch{1.3}
	\caption{Statistics on \emph{Notredame}}
	\label{tab:tab3}
	\begin{tabular}{c|ccccc}
		\hline
		Epoch-$th$ & 30    & 70    & 110   & 150   & 190   \\ \hline
		$\Expect[\theta^r](10^{-1})$       & -3.13 & -3.40 & -3.57 & -3.65 & -3.69 \\ \hline
		$\Std[\theta^r](10^{-1})$       & 2.33  & 2.25  & 2.20  & 2.15  & 2.13  \\ \hline
		$\Expect[\theta^+](10^{-1})$        & 8.26  & 8.14  & 8.06  & 8.02  & 8.01  \\ \hline
		$\Std[\theta^+](10^{-1})$       & 2.18  & 2.09  & 2.01  & 1.96  & 1.95  \\ \hline
		$\Expect[\theta^-]$       & 1.14  & 1.15  & 1.16  & 1.17  & 1.17  \\ \hline
		$\Std[\theta^-](10^{-2})$       & 8.66  & 8.60  & 8.60  & 8.60 & 8.60
		\\	\hline	
		$\Expect[P^+]$       & 281  & 281  & 281  & 281 & 281
		\\ \hline
		$\Expect[P^-]$       & 298  & 297  & 295  & 295 & 294
		\\
		\hline
	\end{tabular}
\end{table}

\subsection{UBC PhotoTour}
UBC PhotoTour~\cite{winder2007learning} is the most widely used dataset for local descriptor learning. It consists of three subsets \emph{Liberty}, \emph{Yosemite} and \emph{Notredame}. The whole dataset contains more than $1.5$M patches and $500$K labels. Deep descriptors are trained on one subset and tested on the other two. In the standard protocol, the test aims to verify $100$K pairs of patches matching or not. We report the false positive rate at 95$\%$ recall (FPR@95) for verification results in Table~\ref{tab:tab1}. Let A-B represent the result trained on A and then tested on B. Compared with the HyNet, our SDGMNet significantly improves YOS-ND and ND-YOS with large margins of $0.15$ and $0.2$. Only on YOS-LIB a small gap exists. Finally, we obtain a gain of $0.08$ on the mean FPR@95. The improvements are considerable for the saturated performance.

\begin{figure*}[t]
	\centering
	\includegraphics[width=0.9\linewidth]{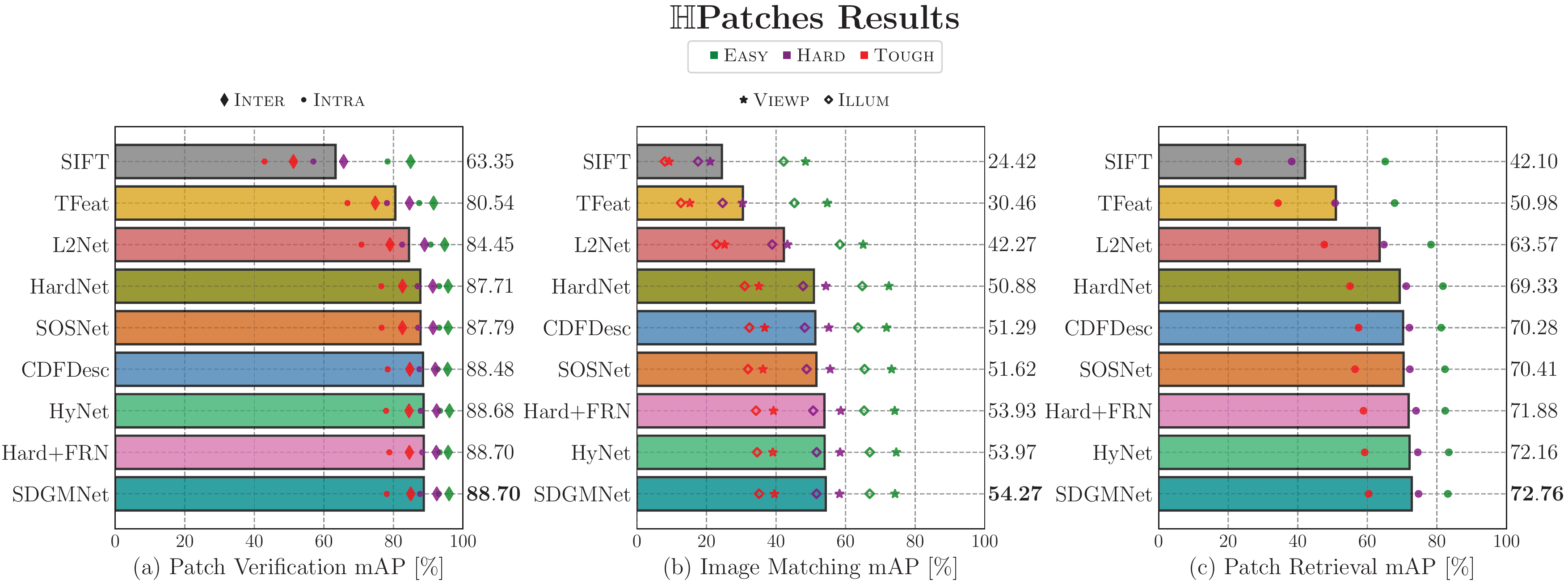}
	\caption{Test on split `a' of HPatches benchmark. All deep descriptors are trained on \emph{Liberty} of PhotoTour. We report mean average precision (mAP) as evaluation metric. Results of subtasks are marked with different colors and patterns. The bars show the mean scores of subtasks. Mean scores are ranked from lowest to highest. }
	\label{fig:fig4}
\end{figure*}

\begin{table}[t]
	\centering
	\setlength{\tabcolsep}{2.5mm}
	\renewcommand\arraystretch{1.3}
	\caption{Statistics on \emph{Yosemite}}
	\label{tab:tab4}
	\begin{tabular}{c|ccccc}
		\hline
		Epoch-$th$ & 30    & 70    & 110   & 150   & 190   \\ \hline
		$\Expect[\theta^r](10^{-1})$       & -3.61 & -3.86 & -4.00 & -4.07 & -4.11 \\ \hline
		$\Std[\theta^r](10^{-1})$       & 2.46  & 2.41  & 2.36  & 2.33  & 2.31  \\ \hline
		$\Expect[\theta^+](10^{-1})$        & 8.10  & 7.97  & 7.89  & 7.85  & 7.84  \\ \hline
		$\Std[\theta^+](10^{-1})$       & 2.36  & 2.30  & 2.24  & 2.21  & 2.19  \\ \hline
		$\Expect[\theta^-]$       & 1.17  & 1.18  & 1.19  & 1.19  & 1.19  \\ \hline
		$\Std[\theta^-](10^{-2})$       & 6.50  & 6.65  & 6.81  & 6.90 & 6.97
		\\	\hline	
		$\Expect[P^+]$       & 284  & 282  & 280  & 280 & 279
		\\ \hline
		$\Expect[P^-]$       & 304  & 301  & 300  & 298 & 297
		\\
		\hline
	\end{tabular}
\end{table}

\begin{figure*}[t]
	\centering
	\subfigure[SIFT]{
	\begin{minipage}[b]{0.32\linewidth}
			\includegraphics[width=1\linewidth]{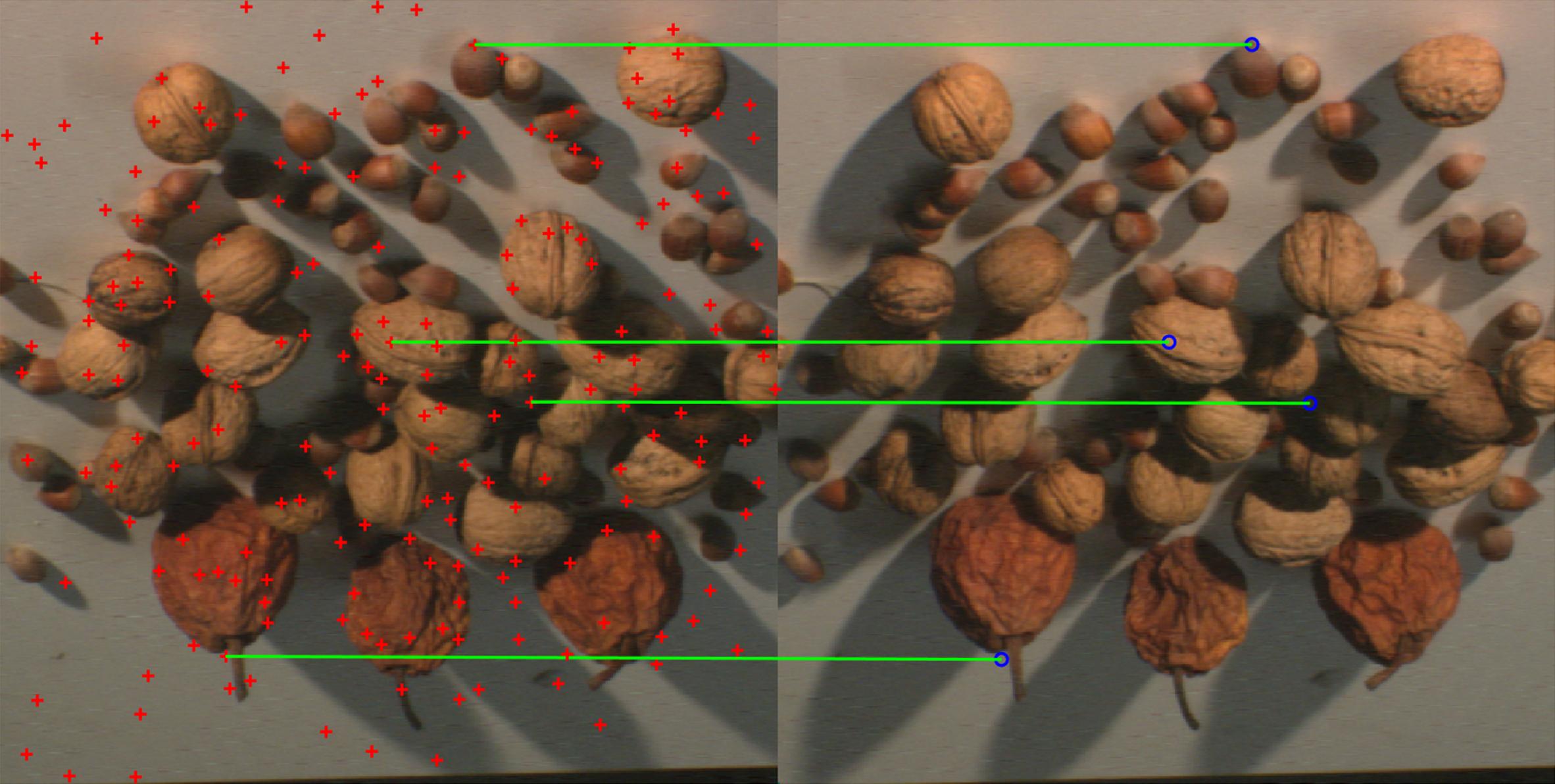}\vspace{4pt}
			\includegraphics[width=1\linewidth]{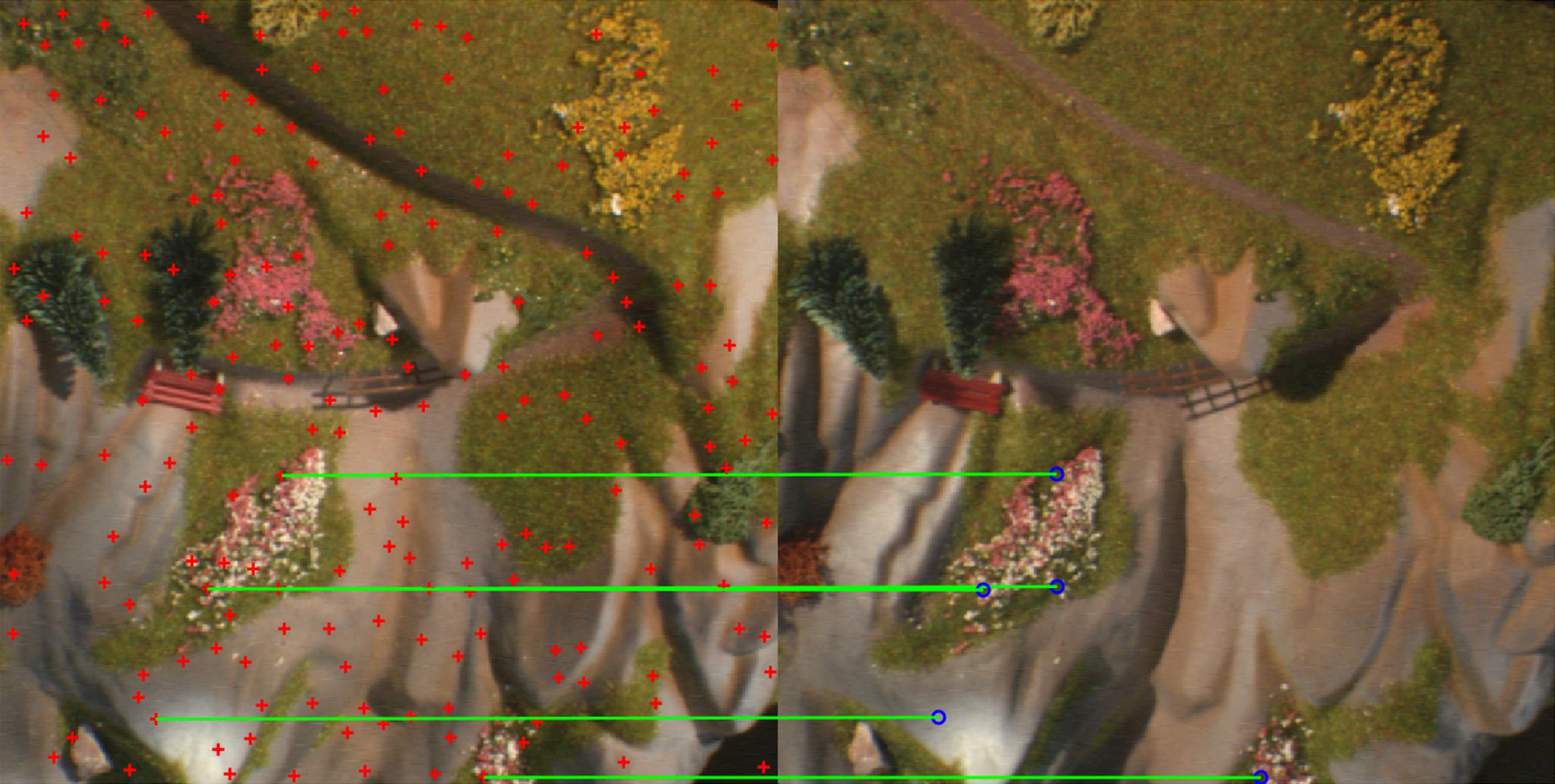}\vspace{4pt}
			\includegraphics[width=1\linewidth]{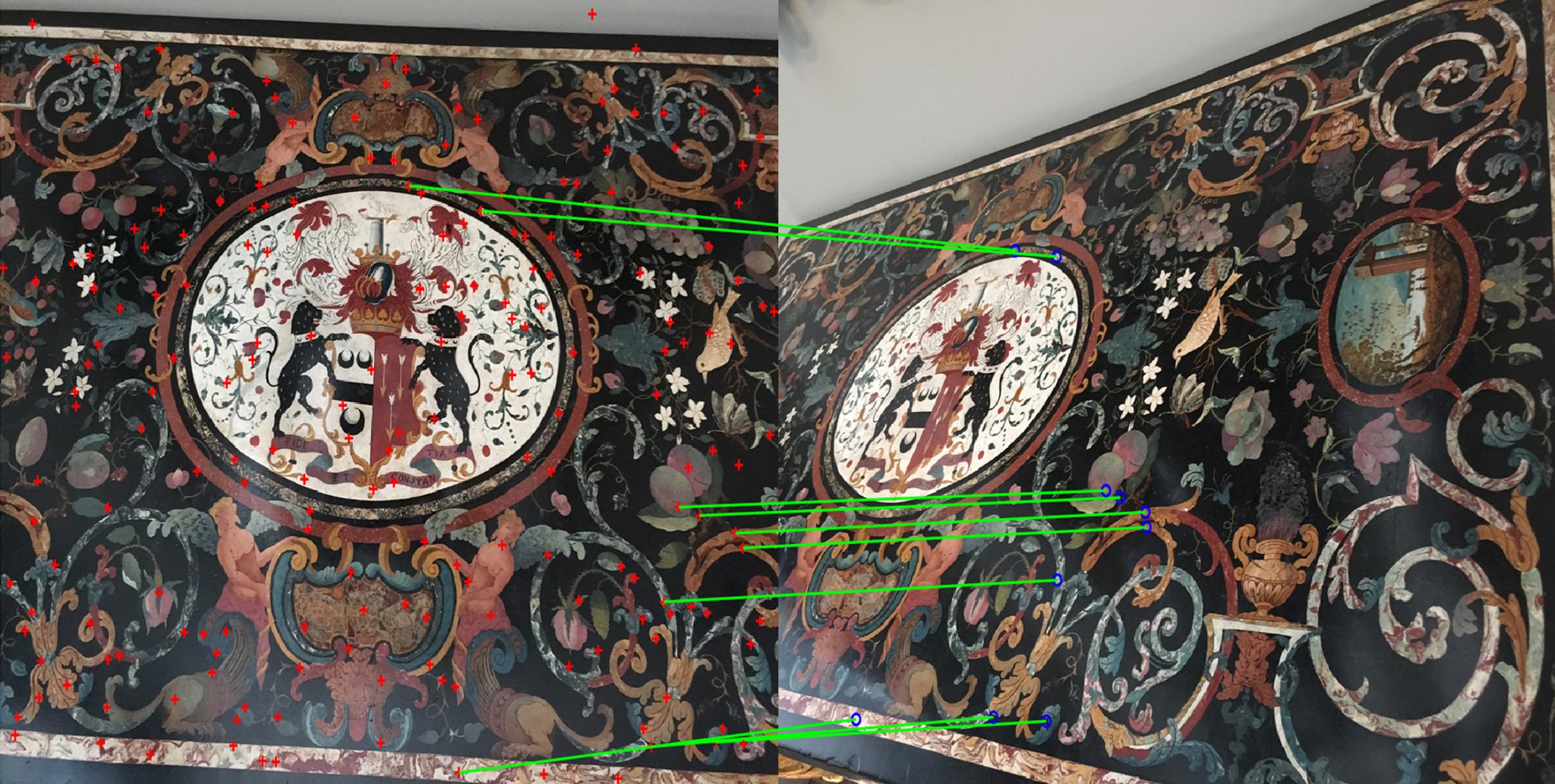}\vspace{4pt}
			\includegraphics[width=1\linewidth]{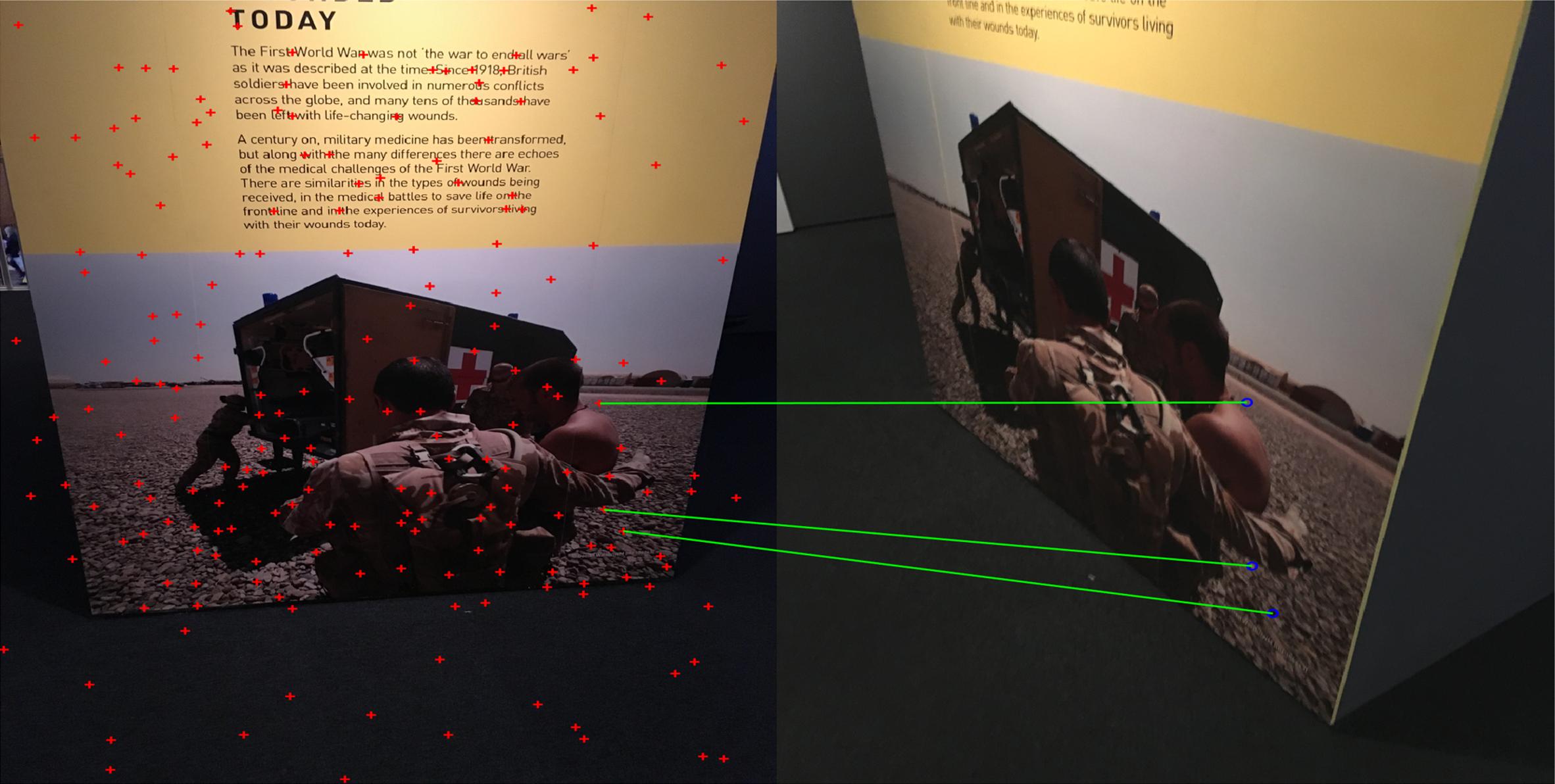}\vspace{4pt}
			\includegraphics[width=1\linewidth]{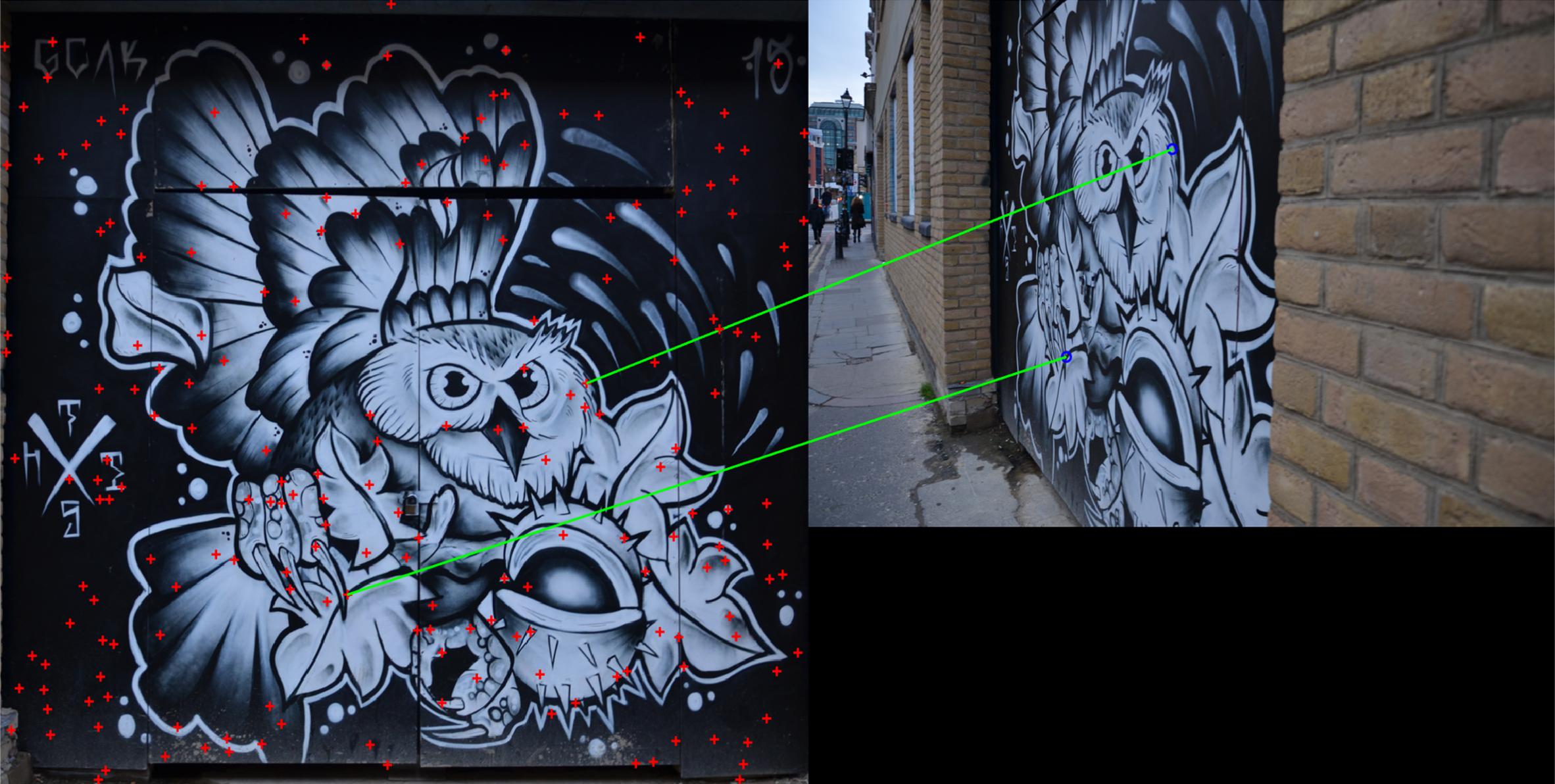}\vspace{4pt}
	\end{minipage}}
	\subfigure[HardNet]{
		\begin{minipage}[b]{0.32\linewidth}
			\includegraphics[width=1\linewidth]{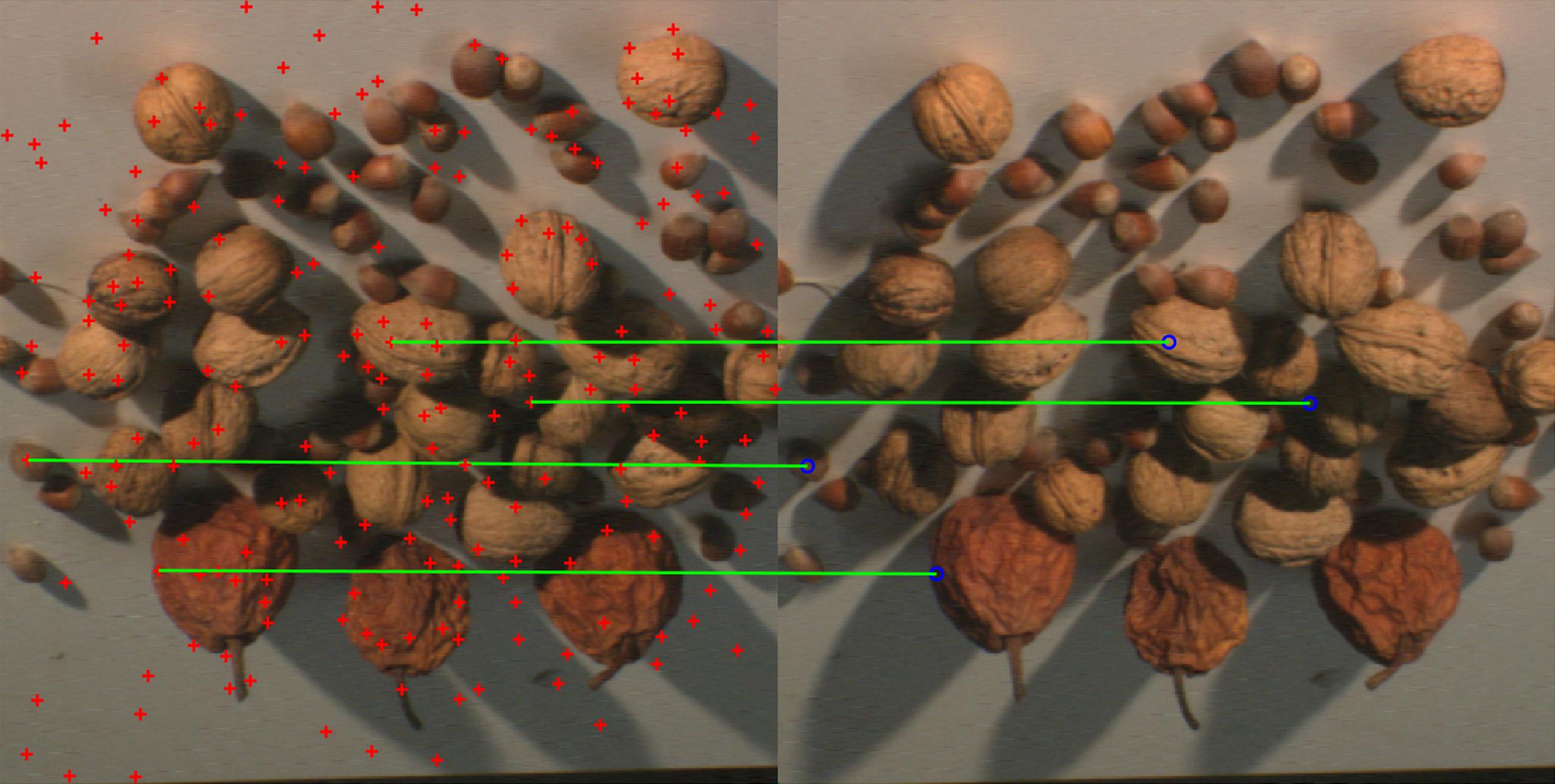}\vspace{4pt}
			\includegraphics[width=1\linewidth]{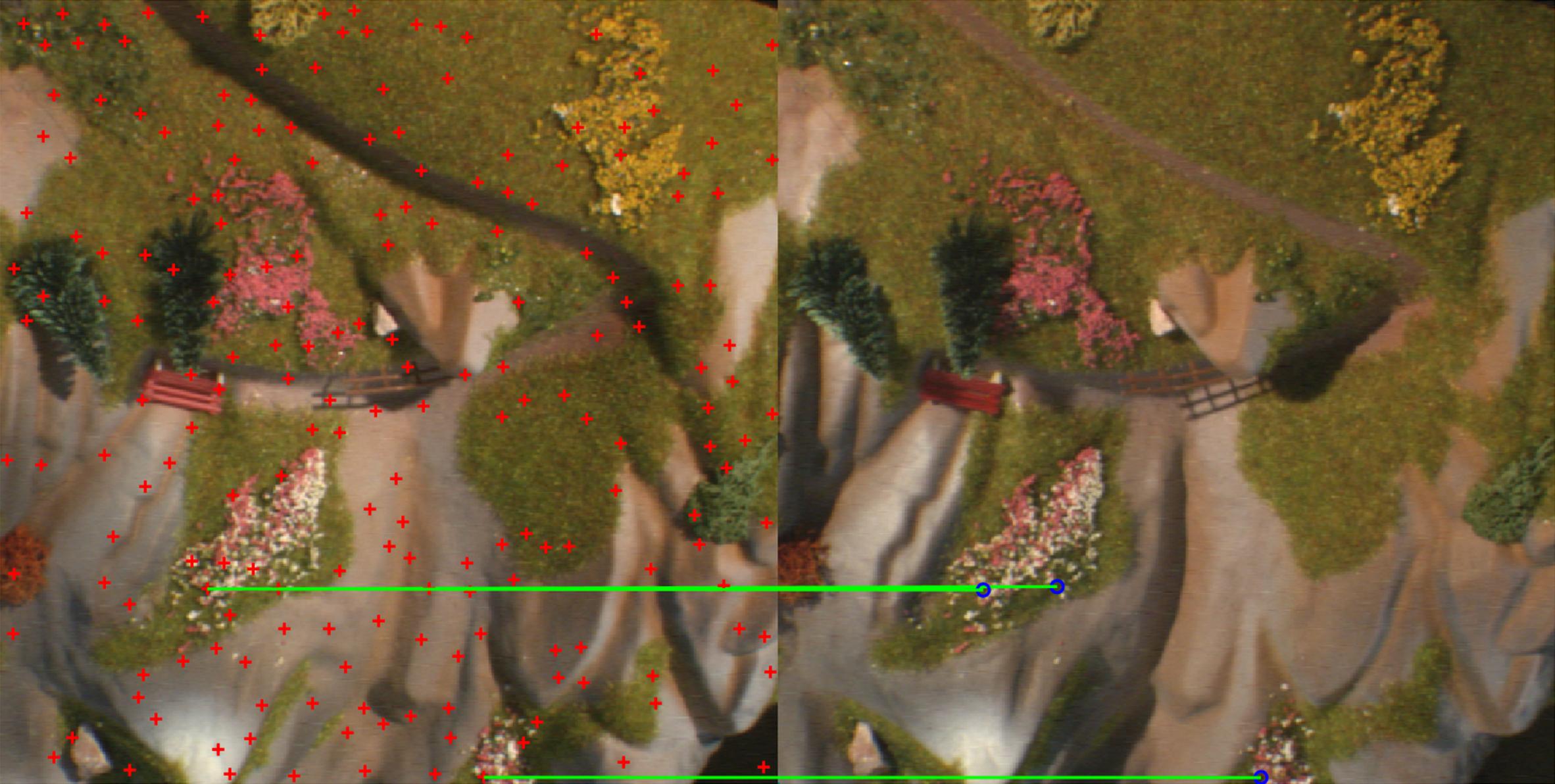}\vspace{4pt}
			\includegraphics[width=1\linewidth]{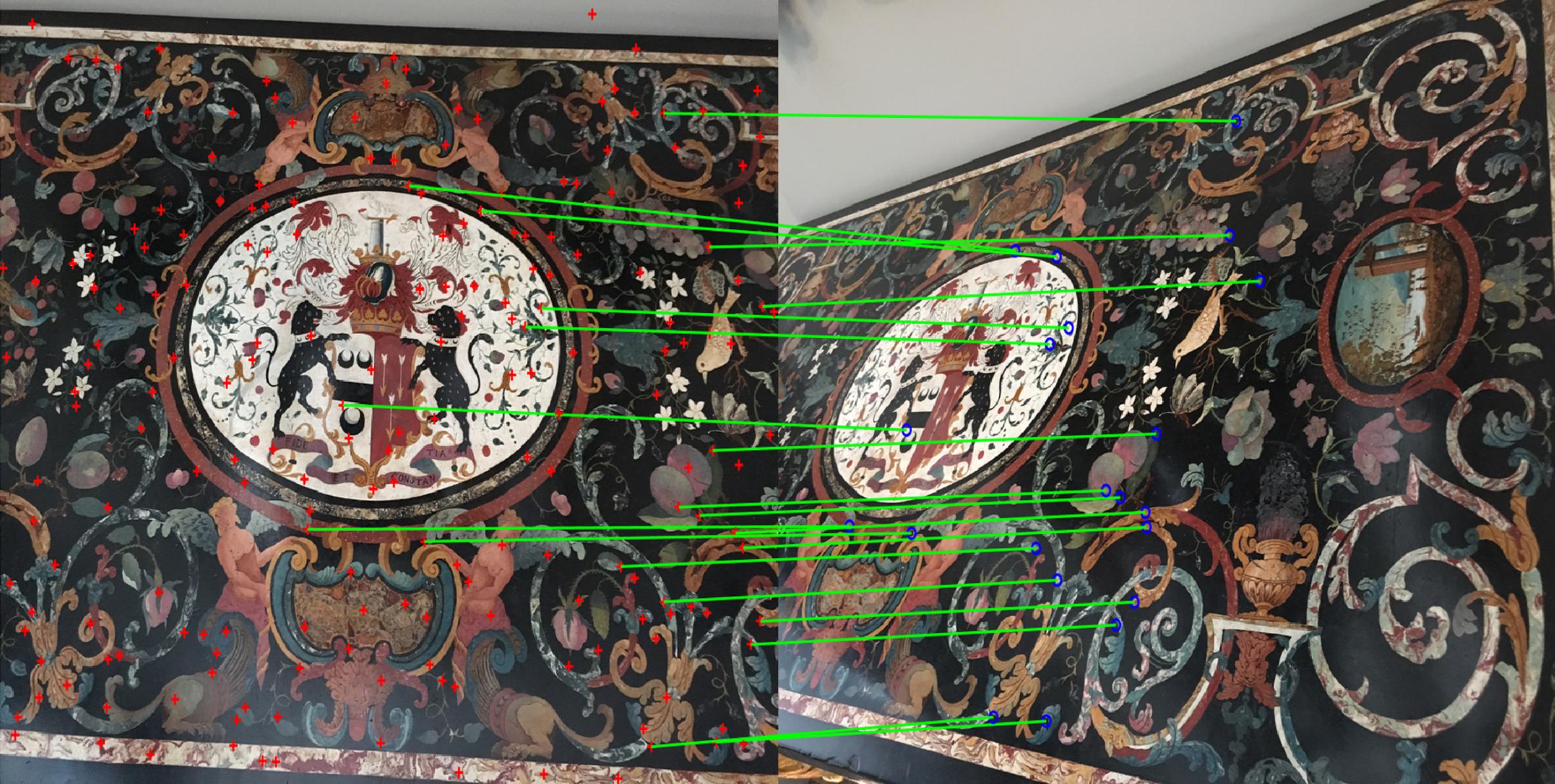}\vspace{4pt}
			\includegraphics[width=1\linewidth]{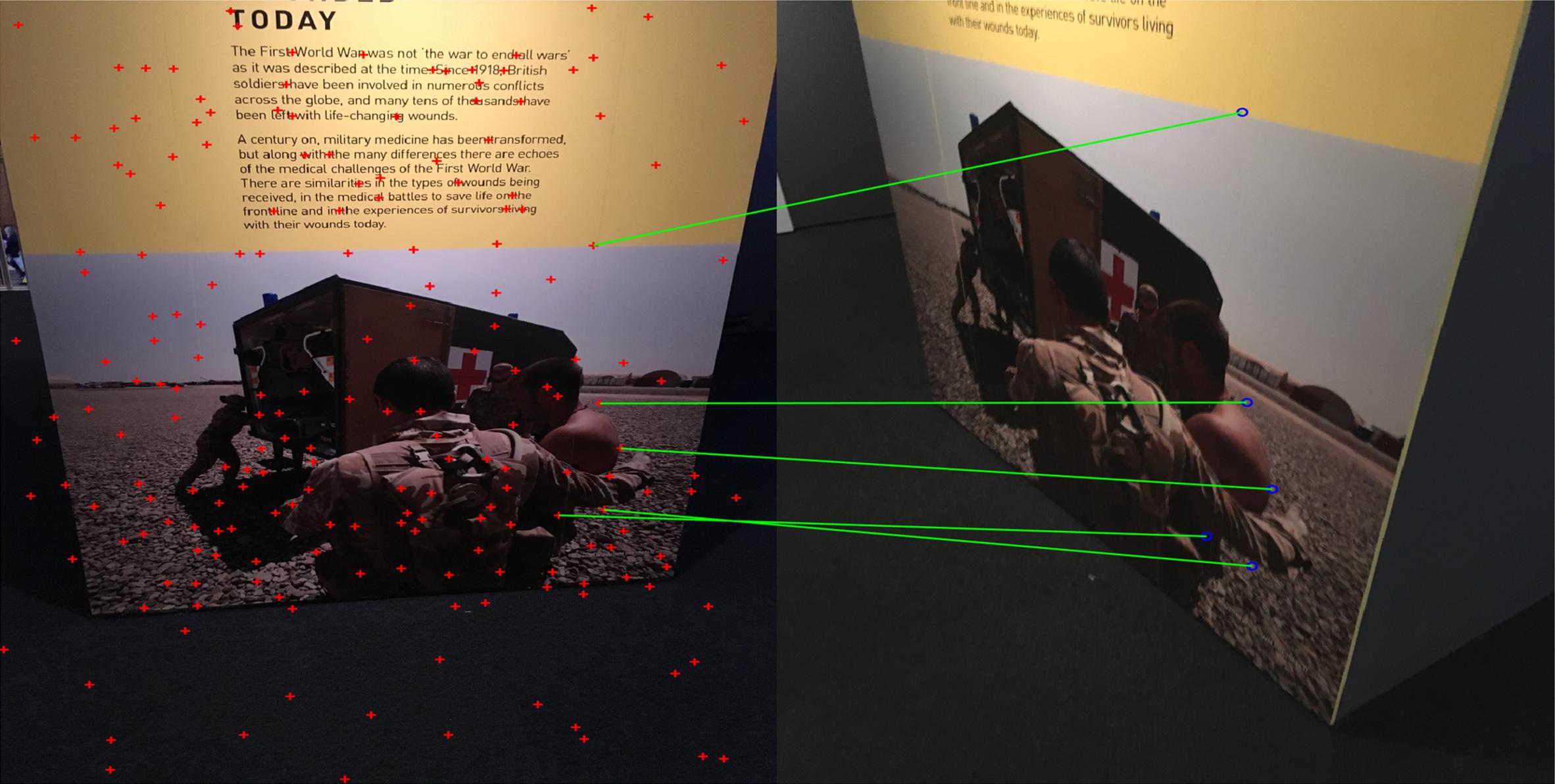}\vspace{4pt}
			\includegraphics[width=1\linewidth]{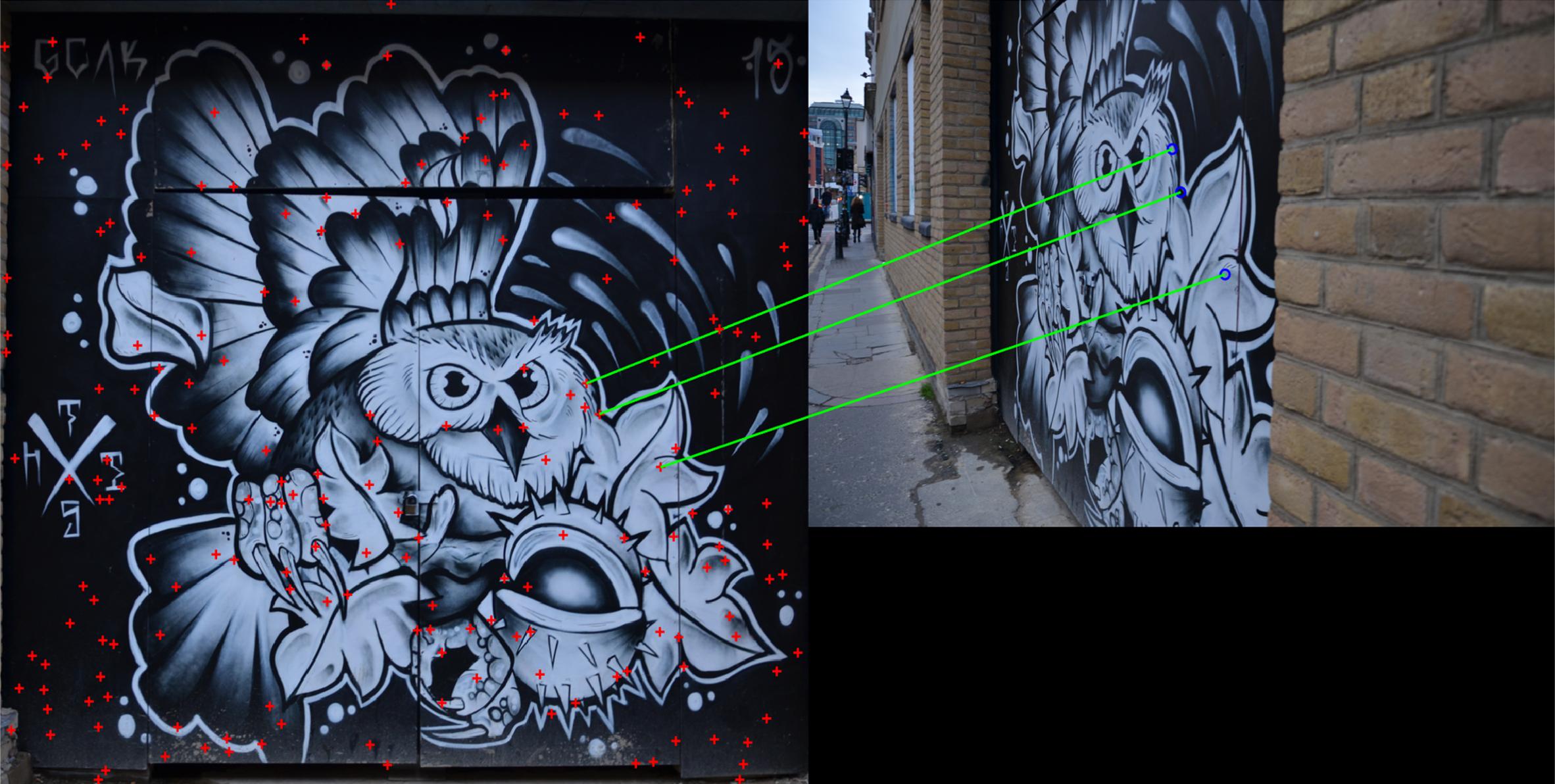}\vspace{4pt}
	\end{minipage}}
	\subfigure[SDGMNet]{
		\begin{minipage}[b]{0.32\linewidth}
			\includegraphics[width=1\linewidth]{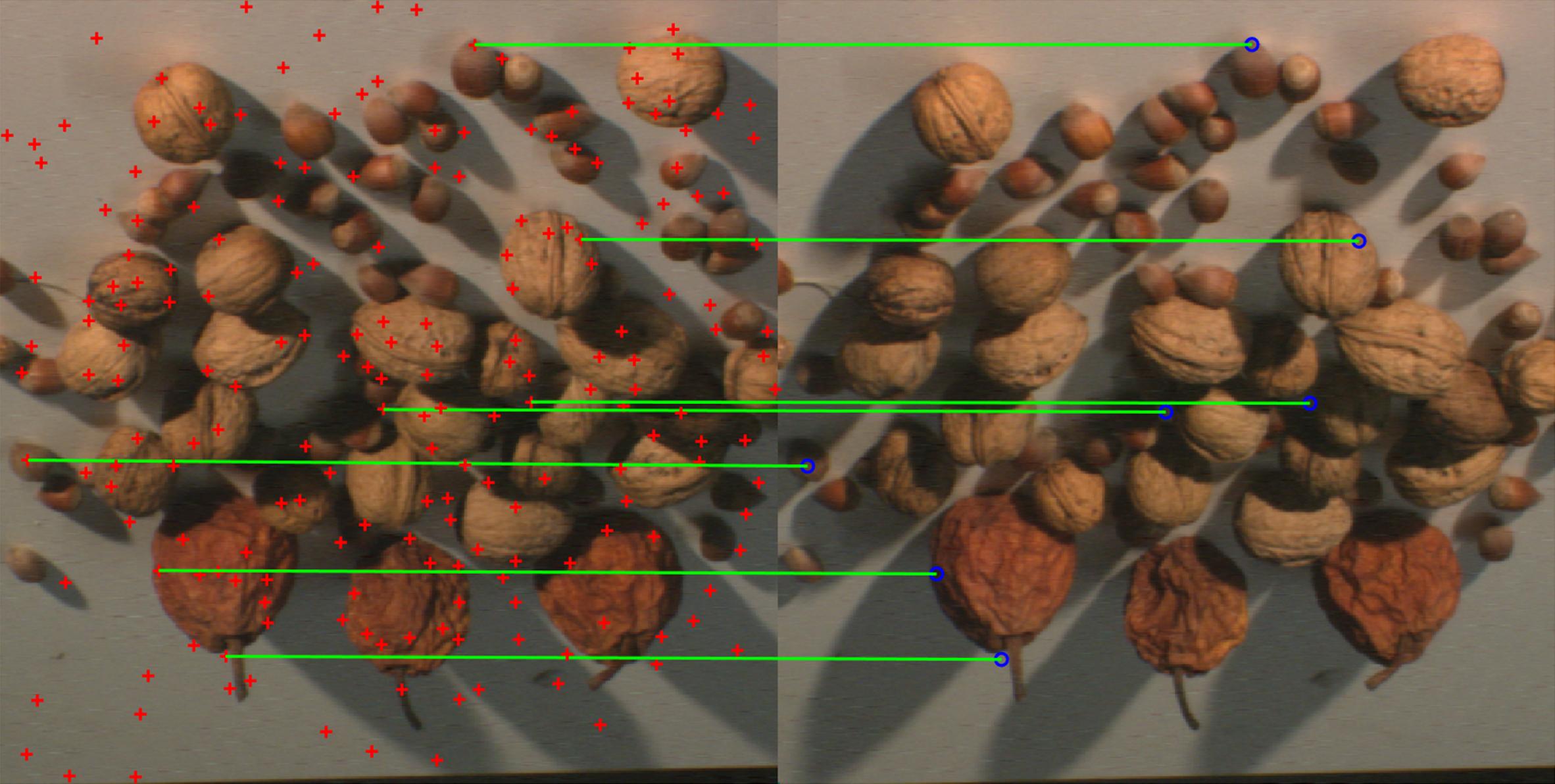}\vspace{4pt}
			\includegraphics[width=1\linewidth]{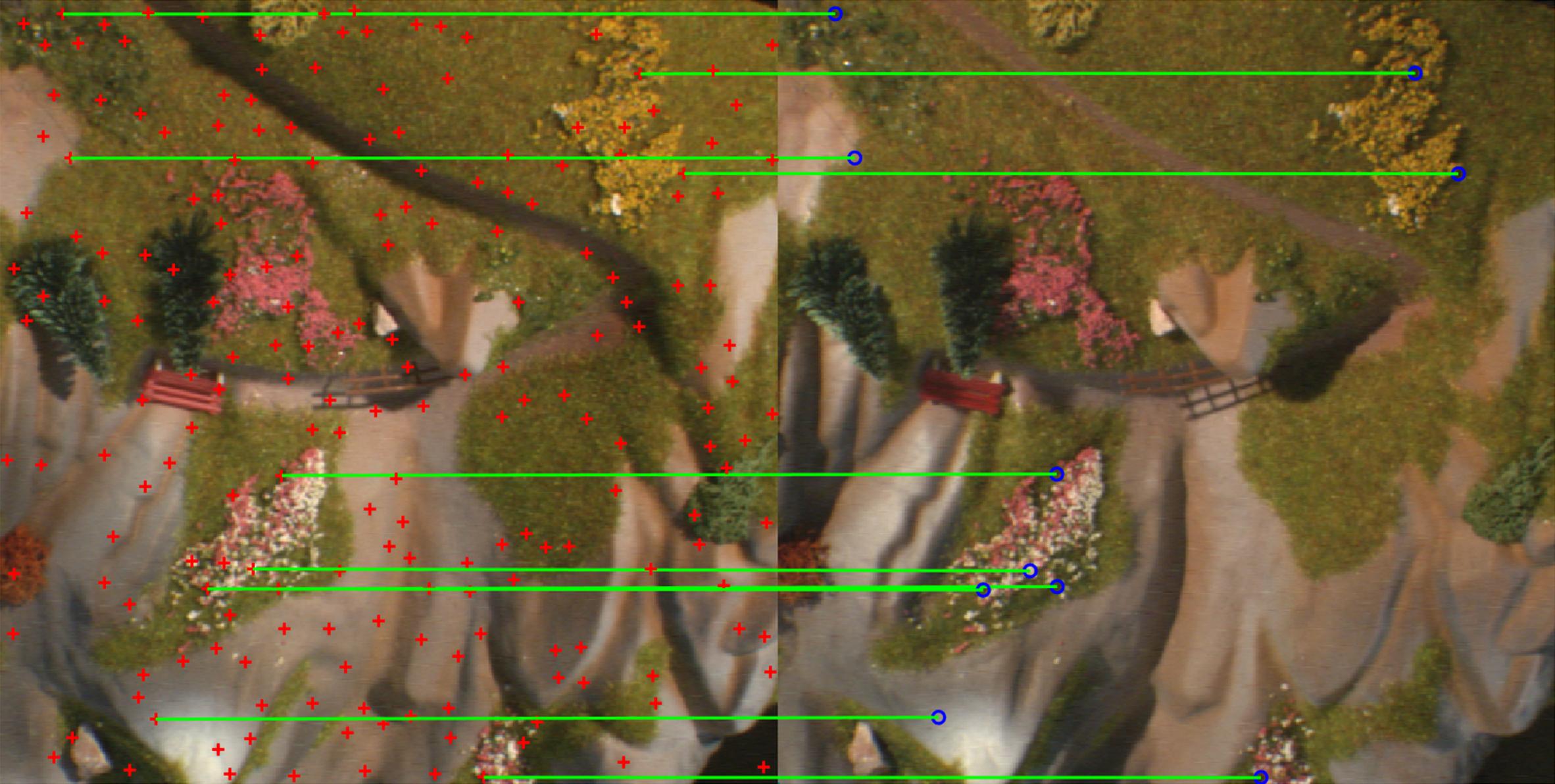}\vspace{4pt}
			\includegraphics[width=1\linewidth]{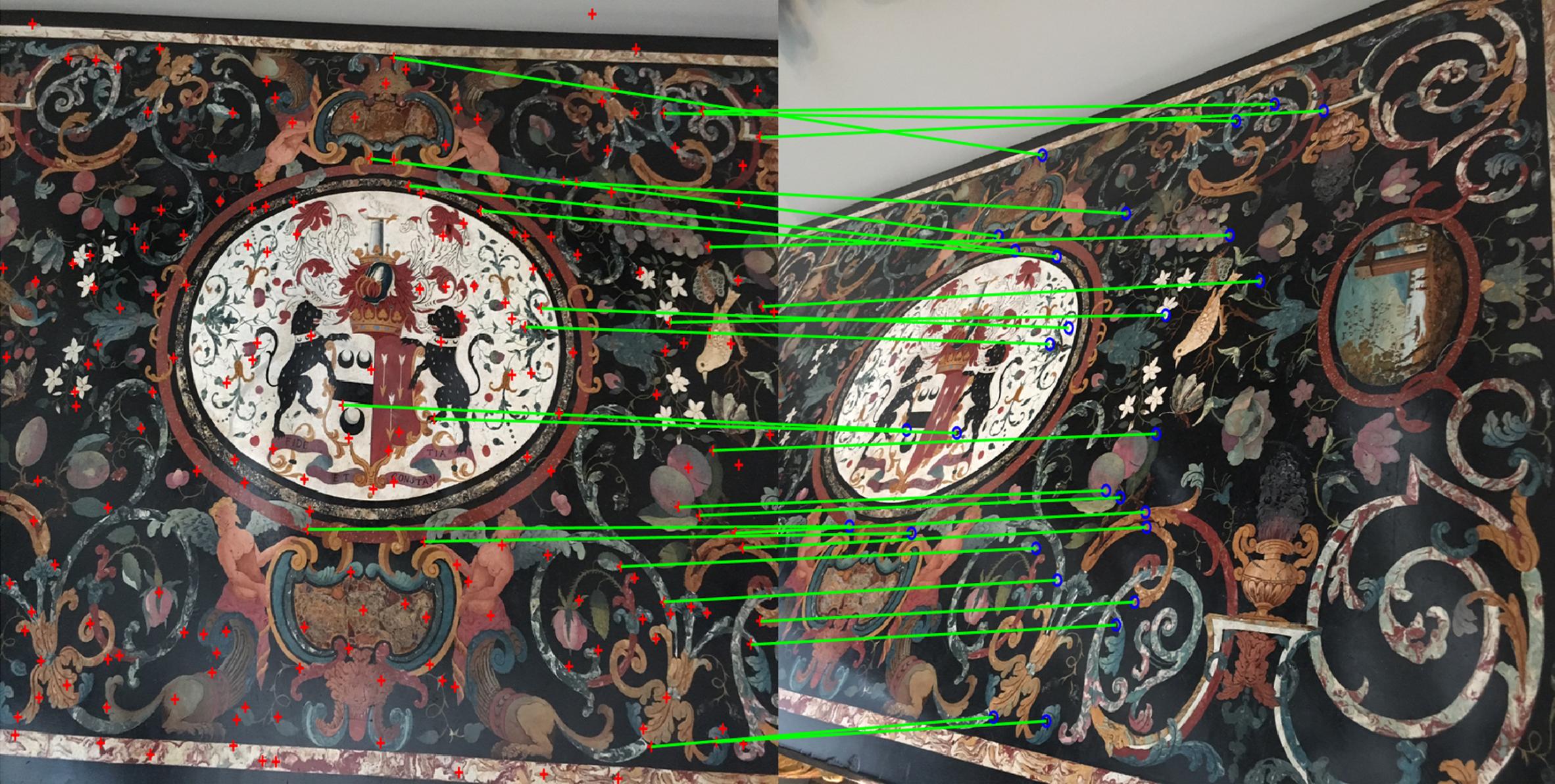}\vspace{4pt}
			\includegraphics[width=1\linewidth]{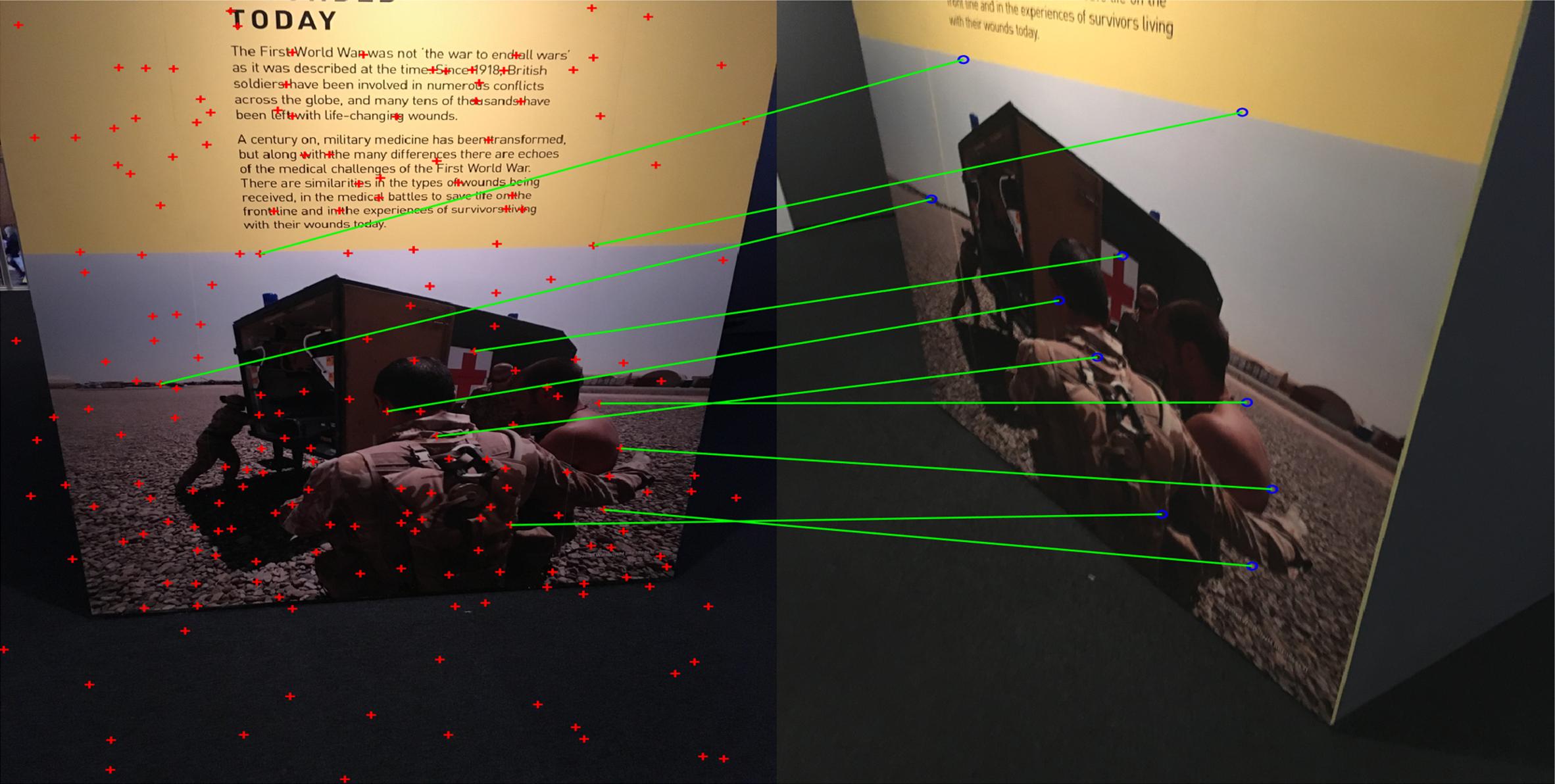}\vspace{4pt}
			\includegraphics[width=1\linewidth]{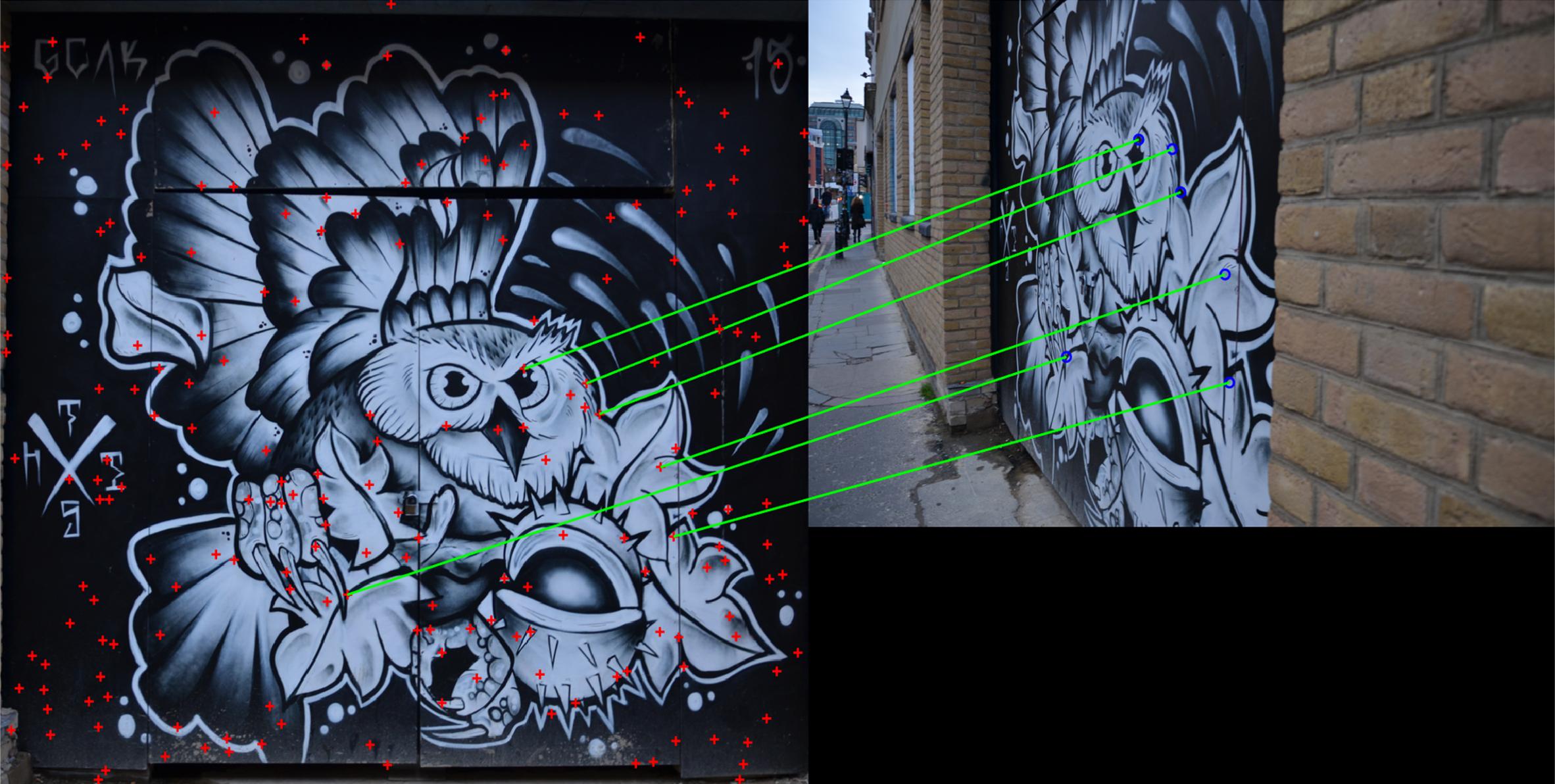}\vspace{4pt}
	\end{minipage}}
	\caption{Feature matching on extreme raw HPatches. The left image of each pair is the reference. The unregistered image is on the right. The fist two rows of pairs suffer from illustration variance. The other pairs are pictured from different viewpoints. Obviously, there exists homography between main regions in each image pair. Keypoints are extracted by Difference-of-Gaussians (DOG) detector~\cite{lowe2004distinctive}. $200$ keypoints on the reference image are randomly selected and marked with red `$+$'. All keypoints on the unregistered image are preserved for matching. Descriptors of keypoints are matched by nearest neighbor principle without ratio test. The correctly matched pairs are linked with green lines, while the others are not shown for clarity. }
	\label{fig:fig5}
\end{figure*}

\begin{table*}[t]
	\centering
	\setlength{\tabcolsep}{5.8mm}
	\renewcommand\arraystretch{1.2}
	\caption{Evaluation on five subsets of ETH 3D reconstruction benchmark. Numbers of unregistered images in different subsets range from 8 to 1463. Five crucial indexes from the benchmark are selected to report. The first and the second best score are marked in red and blue, respectively.}
	\label{tab:tab5}
	\begin{tabular}{lllllll}
		\hline
		&             & \begin{tabular}[c]{@{}l@{}}\#Reg.\\ Images\end{tabular} & \begin{tabular}[c]{@{}l@{}}\#Sparse\\ Points\end{tabular} & \begin{tabular}[c]{@{}l@{}}\#Dense\\ Points\end{tabular} & \begin{tabular}[c]{@{}l@{}}Track\\ Length\end{tabular} & \begin{tabular}[c]{@{}l@{}}Reproj.\\ Error\end{tabular} \\ \hline
		& HardNet & 8                                                       & 8.7K                                                      & 239K                                                     & 4.30                                                   & {\color[HTML]{CB0000} 0.50px}                           \\
		& SOSNet  & 8                                                       & 8.7K                                                      & 239K                                                     & 4.31                            & {\color[HTML]{CB0000} 0.50px}                           \\
		& HyNet   & 8                                                       & {\color[HTML]{3166FF} 8.9K}                               & {\color[HTML]{CB0000} 246K}                              & {\color[HTML]{CB0000} 4.32}                            & 0.52px                                                  \\
		\multirow{-4}{*}{\begin{tabular}[c]{@{}l@{}}Herzjesu\\ (8 images)\end{tabular}}               &SDGMNet  & 8                                                       & {\color[HTML]{CB0000} 9.0K}                               & 233K                                                     & {\color[HTML]{CB0000} 4.32}                            & 0.53px                                                  \\ \hline
		& HardNet & 11                                                      & 16.3K                                                     & 303K                              & 4.91                                                   & 0.47px                                                  \\
		& SOSNet  & 11                                                      & 16.3K                              & {\color[HTML]{CB0000} 306K}                              & 4.92                                                   & {\color[HTML]{CB0000} 0.46px}                           \\
		& HyNet   & 11                                                      & {\color[HTML]{CB0000} 16.5K}                              & 303K                              & {\color[HTML]{3166FF} 4.93}                            & 0.48px                                                  \\
		\multirow{-4}{*}{\begin{tabular}[c]{@{}l@{}}Fountain\\ (11 images)\end{tabular}}               & SDGMNet  & 11                                                      & {\color[HTML]{CB0000} 16.5K}                              & 295K                                                     & {\color[HTML]{CB0000} 4.94}                            & 0.48px                                                  \\ \hline
		& HardNet & 128                                                     & 159K                                                      & 2.12M                             & {\color[HTML]{CB0000} 5.18}                            & {\color[HTML]{CB0000} 0.62px}                           \\
		& SOSNet  & 128                                                     &  160K                               & 2.12M                             & {\color[HTML]{3166FF} 5.17}                            & 0.63px                                                  \\
		& HyNet   & 128                                                     & {\color[HTML]{CB0000} 166K}                               & 2.12M                             & 5.14                                                   & 0.63px                                                  \\
		\multirow{-4}{*}{\begin{tabular}[c]{@{}l@{}}South\\ Building\\ (128 images)\end{tabular}}     & SDGMNet  & 128                                                     & {\color[HTML]{CB0000} 166K}                               & 2.11M                                                    & 5.16                                                   & 0.65px                                                  \\ \hline
		& HardNet & 697                             & 261K                                                      & 1.27M                                                    & 4.16                                                   & 0.98px                                                  \\
		& SOSNet  & 675                                                     & 240K                                                      & 1.27M                                                    & {\color[HTML]{CB0000} 4.40}                            & {\color[HTML]{CB0000} 0.94px}                           \\
		& HyNet   & 697                                                     & {\color[HTML]{CB0000} 337K}                               & 1.25M                                                    & 3.93                                                   & 0.98px                                                  \\
		\multirow{-4}{*}{\begin{tabular}[c]{@{}l@{}}Madrid\\ Metropolis\\ (1334 images)\end{tabular}} & SDGMNet  & {\color[HTML]{CB0000} 830}                              & {\color[HTML]{3166FF} 278K}                               & {\color[HTML]{CB0000} 1.30M}                             &{\color[HTML]{3166FF} 4.18}                                                   &{\color[HTML]{3166FF} 0.96px}                                                  \\ \hline
		& HardNet & 1018                                                    & {\color[HTML]{3166FF} 827K}                               & 2.06M                                                    & 2.56                                                   & 1.09px                                                  \\
		& SOSNet  & 1129                                                    & 729K                                                      &{\color[HTML]{3166FF} 3.05M}                                                    &{\color[HTML]{3166FF} 3.85}                                                   & {\color[HTML]{CB0000} 0.95px}                            \\
		& HyNet   & {\color[HTML]{3166FF} 1181}                             & {\color[HTML]{CB0000} 927K}                               & 2.93M                                                    & 3.49                                                   & 1.05px                                                  \\
		\multirow{-4}{*}{\begin{tabular}[c]{@{}l@{}}Gendar-\\ menmarkt\\ (1463 images)\end{tabular}}  & SDGMNet  & {\color[HTML]{CB0000} 1200}                             & 763K                                                      & {\color[HTML]{CB0000} 3.25M}                             & {\color[HTML]{CB0000} 4.10}                            &{\color[HTML]{3166FF}
			1.02px}\\ \hline
	\end{tabular}
\end{table*}

Moreover, since statistics are the cores of our dynamic modulation, we show all statistics on three subsets at several epochs in Tables~\ref{tab:tab2}, \ref{tab:tab3} and \ref{tab:tab4}. As we can see, statistics are different among training phases and datasets. Especially, $\Expect[\theta^r]$ and $\Expect[\theta^+]$ vary apparently, which make the major contributions on our dynamic gradient modulation. In contrast, $\Expect[P^+]$ and $\Expect[P^-]$ keep stable. It suggests that modulation functions change with data distributions dynamically.

\subsection{HPatches}
HPatches~\cite{balntas2017hpatches} is a more comprehensive benchmark that evaluates descriptors on three tasks: patch verification, image matching and patch retrieval. According to geometric distortion, subtasks are categorized into \emph{Easy}, \emph{Hard} and \emph{Tough}. Furthermore, patch pairs from the same or different image sequences are separated into two test subsets for verification, denoted by \emph{Intra} and \emph{Inter}, respectively. And the matching task is designed to evaluate the viewpoint (\emph{VIEWP}) and illumination (\emph{ILLUM}) invariance of descriptors. For a fair comparison, we train SDGMNet on \emph{Liberty} of UBC PhotoTour as other deep descriptors did. Our SDGMNet surpasses predecessors on all three tasks as shown in Fig.~\ref{fig:fig4}. In image matching task, while there is no gap between Hard+FRN and HyNet architecture, we improve Hard+FRN with a gain $0.34$. In the patch retrieval task, we outperform HyNet with a margin $0.6$. The margin is larger than that between SOSNet and CDF whose encoders are the same. In fact, we employ the pre-trained model reported in Table~\ref{tab:tab1}, which shows the strong generalization of SDGMNet.

However, the evaluation metric, mAP, in standard protocol cannot appropriately reveal the feature matching performance in practice. Because nearest neighbor matching is the most popular principle for feature matching. Neighbors that are not the nearest make different contributions to the mAP, which run counter to the point of nearest neighbor principle. Thus, we further evaluate SIFT, HardNet and our SDGMNet on raw images of HPatches with nearest neighbor matching. Extreme matching results are shown in Fig.~\ref{fig:fig5}. As we can see, although mAP shown in Fig.~\ref{fig:fig4} is considerable, no more than $15\%$ descriptors are correctly matched. Even in some results of SIFT and HardNet, the number of matching pairs is smaller than $4$, which is the minimum requirement to evaluate a homography. Our SDGMNet performs relatively well in these extreme conditions. The number of matching pairs significantly mount in our method.

\begin{figure*}[t]
	\centering
	\includegraphics[width=1.0\linewidth]{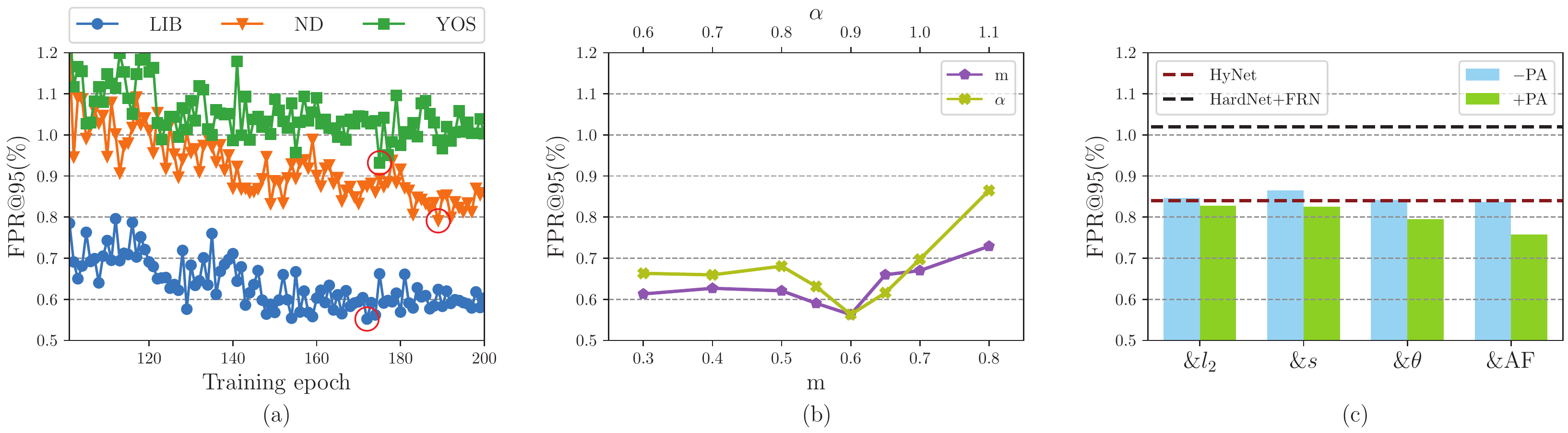}
	\caption{(a) Checkpoints on different subsets of UBC PhotoTour. Each curve illustrates the mean FPR@95 of two test sets. Marked points are the best scores and the correspondings value are recorded in Table~\ref{tab:tab1}. (b) Performances on \emph{Liberty} with different probabilistic margin $m$ or attenuation coefficient $\alpha$. The average of top 10 checkpoints are reported (c) Ablation experiments on full UBC PhotoTour. The red and black dashed line denote the mean FPR@95 of HyNet and HardNet+FRN, respectively.}
	\label{fig:fig6}
\end{figure*}

\subsection{ETH 3D Reconstruction}
ETH benchmark~\cite{schonberger2017comparative} also quantifies descriptors on the image matching task. However, it shows more interest in how the matching performance affects the more challenging and practical 3D reconstruction tasks,~\ie, structure-from-motion (SFM) and Multi-View Stereo (MVS)~\cite{schonberger2015single,schonberger2016structure,schonberger2016pixelwise}. We compare our descriptors with state-of-the-art methods in the standard pipeline of the benchmark. Local patches are extracted by DOG. To directly investigate the performance of descriptors, we abandon the ratio test. All descriptors are trained on \emph{Liberty}. Our pre-trained model reported in Table~\ref{tab:tab1} is applied here again. The results are shown in Table~\ref{tab:tab5}. The number of registered images, reconstructed sparse points, dense points demonstrate the completeness of reconstruction. Mean track length is crucial for the reprojection error and reconstruction accuracy. None of descriptors is absolutely outstanding in relatively small subsets. On Madrid Metropolis, we register the largest number of images, which is $19\%$ more than the second place. Compared with other descriptors, our method gets superb scores on all indexes and balances the completeness and accuracy of reconstruction. Also, results on Gendarmenmarkt confirm the superiority of our method for big datasets.

\begin{figure*}
	\centering
	\includegraphics[width=1.0\linewidth]{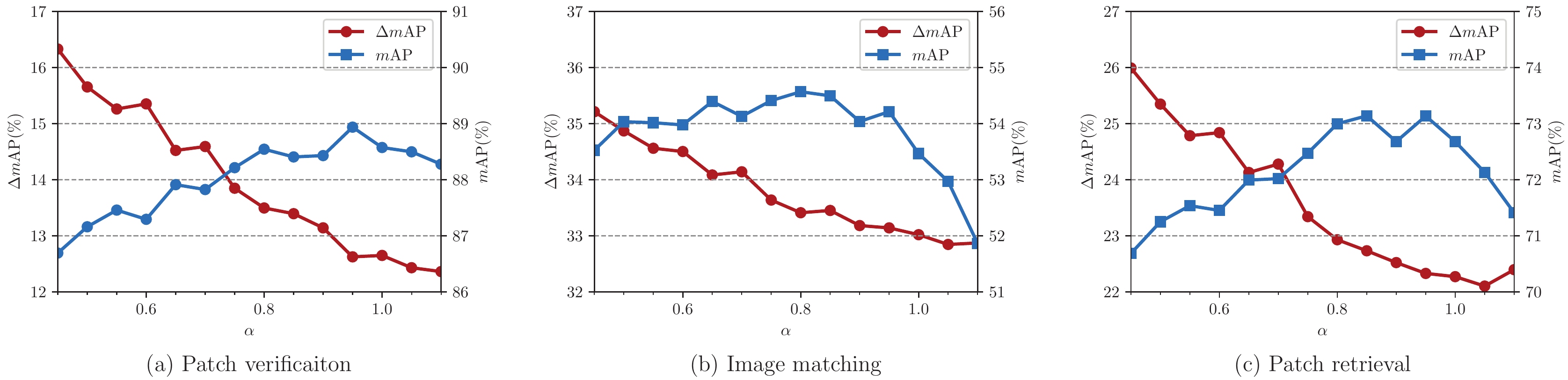}
	\caption{Impact of $\alpha$ on HPatches. Models are trained on $Liberty$. Checkpoints at the last epoch are selected to test on HPatches. $\Delta m$AP denotes the gap between performances on \emph{Tough} and \emph{Easy}. }\label{fig:fig7}
\end{figure*}

\section{Discussion}
\subsection{Reproducibility}
During training on a subset of UBC PhotoTour, we test the model every epoch and report the best core in Table~\ref{tab:tab1} as several works did. Test results from $101$th epoch are shown in Fig.~\ref{fig:fig6} (a). As we can see, the curve of ND easily drops below $0.92$, where $0.92$ are the state-of-the-art before. To be more rigorous, we have trained the model for $10$ times to test the reproducibility. We choice top $10$ checkpoints in every training to provide typical values for the records in Table~\ref{tab:tab1} are $[0.86\pm0.04\text{, }1.55\pm0.14\text{, }0.29\pm0.02\text{, }0.51\pm0.05\text{, }0.88\pm0.08\text{, }0.80\pm0.06\text{, }0.82\pm0.06]$.

\subsection{Impact of $m$ and $\alpha$}
SDGMNet contains two hyperparameters, namely probabilistic hard margin $m$ and attenuation coefficient $\alpha$. To evaluate the impacts of these two hyperparameters, we train SDGMNet on \emph{Liberty} with one of them changing. Then, we report the average of top $10$ checkpoints for higher confidence. The curves of FPR@95 versus $m$ and $\alpha$ are drawn in Fig.~\ref{fig:fig6} (b). As shown, a probabilistic hard margin smaller than $0.6$ degenerates the performance slightly, which demonstrates that probabilistic hard margin can isolate easy triplets more completely than naive CDF-based soft margin. This modification slightly boosts the descriptors, but an excessively large hard margin would release only a few examples into optimization,~\eg., fewer than $30\%$ with $m=0.7$. Too few examples undoubtedly make the model overfit on current batch and weaken the performance. Compared with $\alpha$, varying $m$ leads to smaller fluctuation.Thus, we do not show the joint impact of two hyperparameters in this paper. We leave the discussion of $\alpha$ in Section~\ref{section5.5}.

\subsection{Ablation Study}
We propose four modifications in SDGMNet, including angular distance, auto-focus modulation (AF), probabilistic margin and power adjustment (PA). In fact, we think of angular distance from the perspective of gradient modulation for individual pairs, which shares the same motivation with AF. In other words, AF and implicit modulations in Eqns.~\eqref{eqn:eqn5},~\eqref{eqn:eqn6} and~\eqref{eqn:eqn7} are kinds of self weight, where the term $1/\|\boldsymbol{x}\|$ is omitted. Let \&$\theta$, \&$s$, \&$l_2$ denote self weights computed by Eqns.~\eqref{eqn:eqn5},~\eqref{eqn:eqn6},~\eqref{eqn:eqn7}, respectively. And \&$\text{AF}$ represents our formulation. We assess frameworks that combine different kinds of self weight with probabilistic-margin-based coupled weight. To test the efficiency of PA, we embed PA to those raw frameworks. Raw frameworks are labeled with $-$PA and equipped ones with +PA. Related experiments are conducted on full UBC PhotoTour. The performances are shown in Fig.~\ref{fig:fig6}~(c).

Without PA, all frameworks outperform the HardNet~\cite{movshovitz2017no} embedded with FRN (HardNet+FRN). Their scores have been floating near the previous record of HyNet~\cite{tian2020hynet}. It is worth mentioning that \&$l_2-$PA is equivalent to the HardNet+FRN upgraded with probabilistic margin. Probabilistic margin improves HardNet+FRN by a gain about 0.18. In comparison, a CDF-based soft margin,~\ie, CDFDesc~\cite{zhang2019learning}, improves the HardNet by 0.13 as shown in Table~\ref{tab:tab1}. While the performance is almost saturated, our novel probabilistic margin still generates a larger improvement. However, angular distance and AF reveal only a little distinction without PA. After PA is equipped, all raw frameworks advance. In such circumstance, \&$\theta$+PA and \&$\text{AF}$+PA show their superiority. Especially, \&$\text{AF}$+PA builds up a big lead. These outcomes suggest the efficiencies of angular distance, AF and PA.

\subsection{Analysis on the Ratio of Powers}
\label{section5.5}
Attenuation coefficient $\alpha$ controls the ratio of the normalized positive power to the negative. The impact of $\alpha$ shown in Figs.~\ref{fig:fig6} (b) and \ref{fig:fig7} confirms that an inductive bias to the negative class,~\ie, an attenuation coefficient less than $1$, is helpful for improvement. Moreover, four raw frameworks ($-$PA) in Fig.~\ref{fig:fig6} (c) take different modulations,~\ie, self weights. They hold mean ratios of powers on three subsets as $[1.02\text{, }0.94\text{, }1.00\text{, }0.95]$. Better performances emerge when the ratios are forced to $0.9$. It means that power adjustments facilitate different modulations. Furthermore, results shown in Fig.~\ref{fig:fig4} are mean scores of full UBC PhotoTour, which imply that the impact of ratio takes effect on various training sets.

To give an insight into how the ratio of power affects the performance, we train the models with different $\alpha$ on $\emph{Liberty}$ and test them on HPatches. We introduce a naive metric $\Delta m$AP, which is computed as $m$AP on \emph{Easy} minus $m$AP on \emph{Tough}. As shown in Fig.~\ref{fig:fig7}, while the $m$AP peaks near $\alpha=0.9$ in three tasks, the $\Delta m$AP tends to decline along with the increasing $\alpha$. It can be explained that a ratio small than $1$ leads to a preference on the negative part. The model trained with a small ratio cannot learn the extreme variance of positive pairs. It wrongly regards those hard positive pairs as negative ones. Meanwhile, a sizable ratio is enough to help the model handle those easy positive examples. So the performances on easy examples remain relatively stable. In conclusion, the best ratio depends on the difficulty level of the task. Since the difficulty level can be quantified in image matching task, power adjustment for local descriptor learning will be helpful in practice.

\section{Conclusion}
In this paper, we propose a statistic-based dynamic gradient modulation for local descriptor learning, called SDGMNet. SDGMNet devotes to dynamically rescaling the gradients of pairs. Firstly, SDGMNet conducts deep analysis on back propagation and chooses included angle for distance measure. The angular distance is unbiased for every pair in theory. Secondly, auto-focus modulation is applied to modulate the gradients of individual pairs. It neutralizes the HEM and extreme example suppression according to statistical characteristics of individual pairs. Thirdly, SDGMNet enrolls statistic-based probabilistic margin to modulate the gradients of Siamese pairs,~\ie, triplets. It combines hard and soft HEM on triplets to help stochastic gradient descent optimization converge. Finally, total weights,~\ie, powers of two kinds of pairs are adjusted by gradient normalization and attenuation coefficient. SDGMNet fulfills the theme of modulating gradients dynamically with systematical analysis. Local descriptors extracted by SDGMNet show superiority on various tasks and datasets. Every modification in SDGMNet proves efficient by extensive experiments.

Moreover, our success confirms that some extra problems deserve attentions, including hidden modulation in deep back propagation, strict or flexible HEM, balance between powers. Many metric learning tasks,~\eg, face recognition share similar frameworks with local descriptor learning. They would suffer from the same problems. Local descriptor learning is a hard task featured with few-shot, open-set and large-scale. Since SDGMNet can systematically tackle those problems for the task, it is promising to transplant solutions of SDGMNet into other metric learning tasks. Moreover, modulating gradient may not lead to a significant improvement as changing a model does. But as a crucial part of optimization for deep learning, it still merits further study.

{\small
	\bibliographystyle{plain}
	\bibliography{TIP}
}

\end{document}